%% file: main.tex
\makeatletter\providecommand{\@LN}[2]{}\providecommand{\@LN@col}[1]{}\makeatother
\RequirePackage{fix-cm}
\PassOptionsToPackage{table}{xcolor}
\documentclass[11pt,letterpaper,shortlabels]{berkeley}
\usepackage[authoryear,round]{natbib}
\usepackage{mathtools}
\usepackage{multirow,makecell,array,tabularx}
\usepackage{xspace}
\usepackage{wrapfig}
\usepackage{currfile}
\usepackage{tikz}
\usetikzlibrary{arrows.meta,positioning,fit,calc,backgrounds}
\usepackage{fontawesome5}
\usepackage{algorithm}
\usepackage{algpseudocode}
\usepackage[capitalize]{cleveref}

\definecolor{NDblue}{RGB}{23,50,77}
\definecolor{NDgold}{RGB}{197,139,34}
\hypersetup{colorlinks=true,linkcolor=NDblue,citecolor=NDgold,urlcolor=NDblue,
  pdftitle={What a Calibrated Probability Hides: Belief-Level Post-Training for Decision Models},
  pdfauthor={Xinye Li, Zirui Song, Huaxing Liu, Xiang Wang, Shuai Li, Zixiang Xu, Yinjun Liu, Zhewen Zhang, Xiaojun Chang, Xiuying Chen}}
\renewcommand{\titlefont}{\color{BerkeleyBlue}\normalfont\bfseries
  \fontsize{17.5}{20}\selectfont}
\let\cite\citep   % numeric-style \cite in the source -> (Author, Year)
\definecolor{cvprblue}{rgb}{0.21,0.49,0.74}

\DeclareMathOperator{\clip}{clip}

\newcommand{\KL}{\mathrm{KL}}
\newcommand{\Dobs}{\mathcal{D}^{\mathrm{obs}}}
\newcommand{\Deval}{\mathcal{D}^{\mathrm{eval}}}
\newcommand{\thetaold}{\theta_{\mathrm{old}}}
\makeatletter
\DeclareRobustCommand\onedot{\futurelet\@let@token\@onedot}
\def\@onedot{\ifx\@let@token.\else.\null\fi\xspace}

\makeatother

\title{What a Calibrated Probability Hides:\\Belief-Level Post-Training for Decision Models}
\input{frontmatter}
\correspondingauthor{Xiuying~Chen (\href{mailto:xiuying.chen@mbzuai.ac.ae}{xiuying.chen@mbzuai.ac.ae})}
\begin{abstract}
\input{sec/0_abstract}
\end{abstract}
\begin{document}
\raggedbottom
\maketitle
\suppressfloats[t]  % keep Figure 1 below the title block
\input{sec/body}

\bibliography{main}
\clearpage
\appendix
\input{sec/supp_method}
\end{document}

%% file: frontmatter.tex
\newcommand{\paperinstitutionsymbol}[1]{\textcolor{NDblue}{#1}}
\newcommand{\paperinternmark}{\textcolor{NDgold}{*}}
\newcommand{\papercontactmark}{\textcolor{NDgold}{\ensuremath{\dagger}}}
\newcommand{\paperauthor}[2]{%
  \mbox{#2\textsuperscript{\paperinstitutionsymbol{#1}}}}
\newcommand{\paperaffil}[2]{%
  \mbox{\textsuperscript{\paperinstitutionsymbol{#1}}\,#2}}
\newcommand{\papertitlerule}{%
  {\color{NDgold}\rule{26pt}{0.6pt}}%
  {\color{NDblue!28}\rule{\dimexpr\linewidth-26pt\relax}{0.6pt}}}
\makeatletter
\renewcommand{\@author}{%
  {\normalfont\fontsize{11}{14}\selectfont
   \paperauthor{1,2}{Xinye~Li}\textsuperscript{\paperinternmark}\hspace{0.7em}%
   \paperauthor{1,2}{Zirui~Song}\textsuperscript{\paperinternmark}\hspace{0.7em}%
   \paperauthor{2}{Huaxing~Liu}\hspace{0.7em}%
   \paperauthor{2}{Xiang~Wang}\hspace{0.7em}%
   \paperauthor{2}{Shuai~Li}\hspace{0.7em}%
   \paperauthor{4}{Zixiang~Xu}\\
   \paperauthor{1}{Yinjun~Liu}\hspace{0.7em}%
   \paperauthor{3}{Zhewen~Zhang}\hspace{0.7em}%
   \paperauthor{1}{Xiaojun~Chang}\hspace{0.7em}%
   \paperauthor{1}{Xiuying~Chen}\textsuperscript{\papercontactmark}\par}
  \vspace{6pt}%
  {\normalfont\fontsize{10.5}{14}\selectfont
   \paperaffil{1}{Mohamed bin Zayed University of Artificial Intelligence}\hspace{1.2em}%
   \paperaffil{2}{Alibaba Group}\\
   \paperaffil{3}{Northeastern University}\hspace{1.2em}%
   \paperaffil{4}{University of Southern California}\par}}
\renewcommand{\maketitle}{%
  \begingroup
  \setlength{\parindent}{0pt}%
  \setlength{\parskip}{0pt}%
  \centering
  \vspace*{5pt}%
  {\titlefont\@title\par}
  \vspace{11pt}%
  \@author
  \vspace{8pt}%
  \papertitlerule\par
  \vspace{10pt}%
  \endgroup
  {\abscontent}%
  \thispagestyle{firststyle}}
\makeatother
\fancypagestyle{firststyle}{%
  \fancyhf{}%
  \fancyhead[L]{\raisebox{-0.3\height}{\includegraphics[height=44pt]{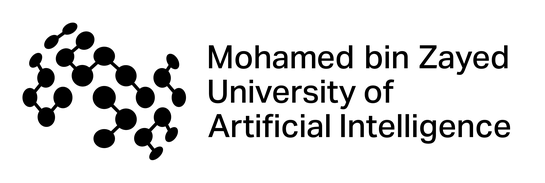}}%
    \hspace{16pt}%
    \raisebox{-0.3\height}{\includegraphics[height=44pt]{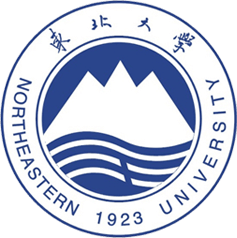}}%
    \hspace{16pt}%
    \raisebox{-0.1\height}{\includegraphics[height=17pt]{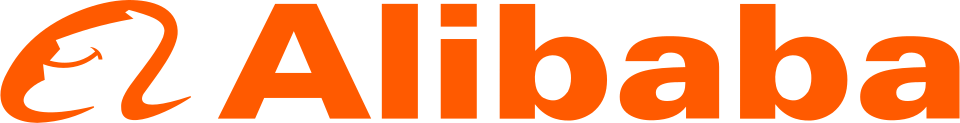}}%
    \hspace{16pt}%
    \raisebox{-0.3\height}{\includegraphics[height=44pt]{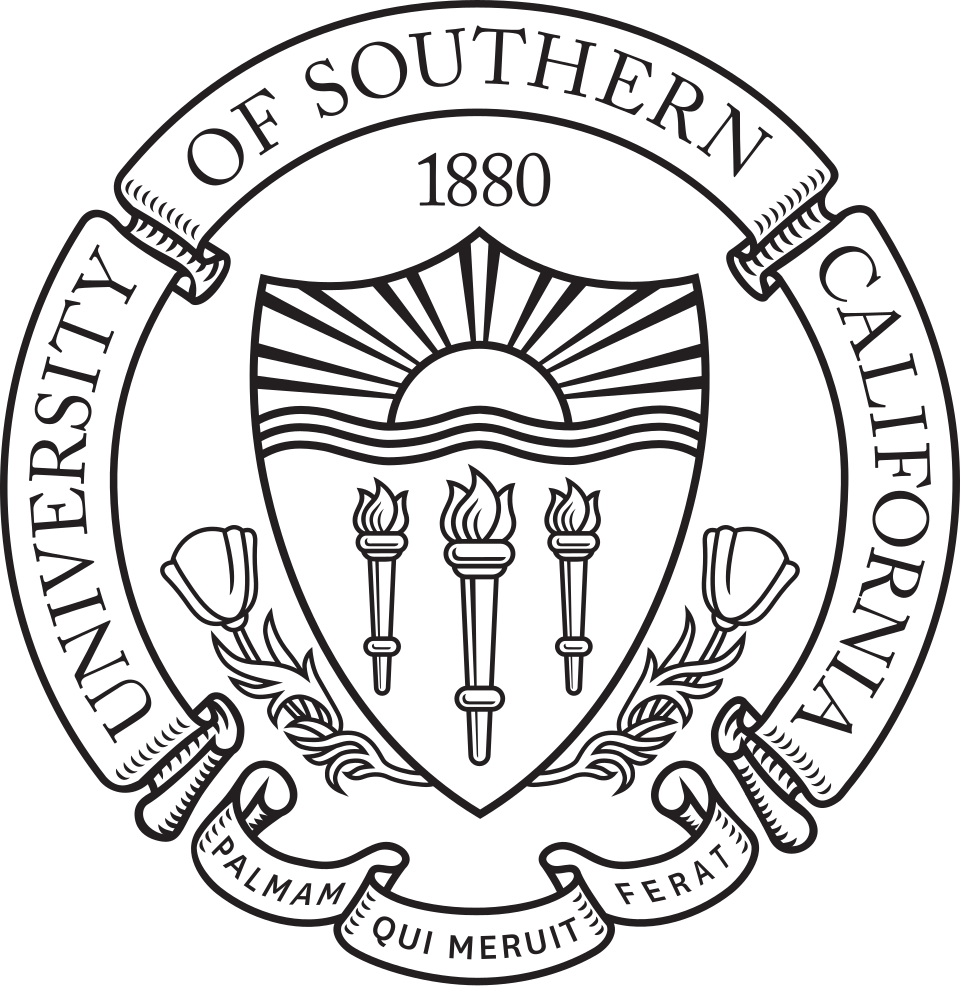}}}%
  \fancyfoot[L]{\normalfont\fontsize{9}{11}\selectfont
    \textsuperscript{\paperinternmark}\,Work done during internship at AMAP, Alibaba Group.\\
    \textsuperscript{\papercontactmark}\,Corresponding author:
    \the\correspondingauthor}}

%% file: sec/0_abstract.tex
A calibrated decision model can predict how often each answer is chosen. The same prediction can reflect disagreement among people who are confident in their choices or uncertainty among people who share a belief. Disagreement calls for identifying whose perspective applies, while uncertainty calls for gathering more evidence. Standard post-training scores a single prediction against one observed choice at a time, learning the pooled distribution without distinguishing these two sources of uncertainty.
We introduce Belief-Grouped Policy Optimization (BGPO). The policy proposes a list of beliefs, estimates their prevalence, and predicts a choice distribution for each. During training, BGPO uses Bayes’ rule to estimate how likely each belief was to generate an observed choice. These estimates weight the choice’s contribution to each belief’s log-score reward. To discover rare beliefs, the policy also proposes lists after seeing a few observed choices and retains them only if they predict held-out choices well. At inference, the policy uses only the question and combines the belief-specific distributions by prevalence to produce a calibrated aggregate prediction.
On benchmarks covering opinions, conversation safety, emotions elicited by artwork, and photo aesthetics, BGPO achieves the lowest conditional, worst-group, and overall Jensen–Shannon divergence across model initializations. It improves conditional divergence on both text and image tasks. On a synthetic image task with known beliefs, BGPO recovers the underlying groups more accurately than baselines trained with belief labels.

%% file: sec/body.tex
\section{Introduction}
\label{sec:intro}

  % --- Para 1: calibration is not enough ---------------------

\begin{figure}[t]
    \centering
    \includegraphics[width=0.72\linewidth]{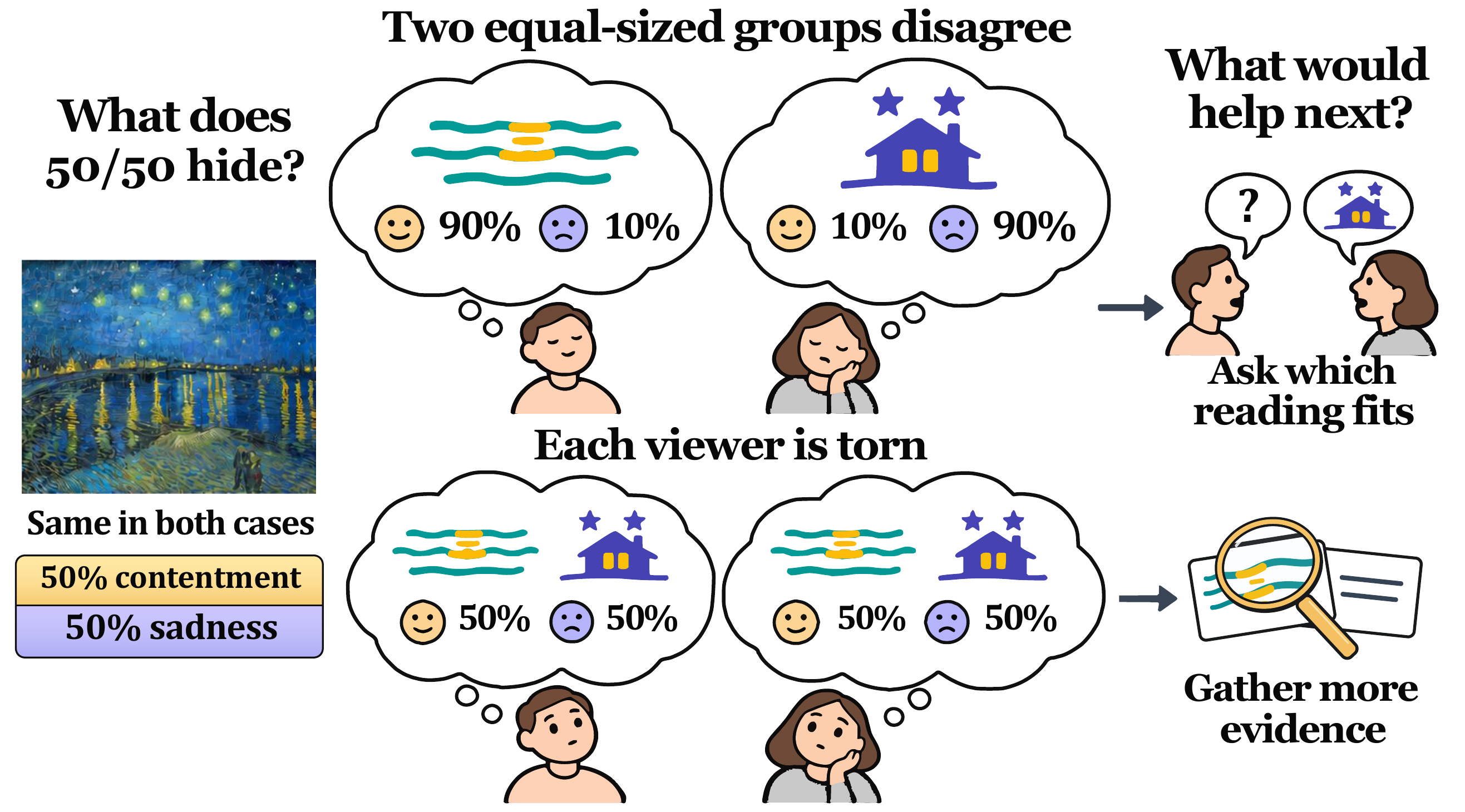}
\caption{\textbf{Two sources of uncertainty behind the same 50/50 prediction.}
\textbf{Top:} Two equally sized groups favor different emotions, so identifying which interpretation fits a viewer helps predict their response.
\textbf{Bottom:} Each viewer is equally torn between contentment and sadness, so gathering more evidence may help resolve their uncertainty.}
    \label{fig:motivation}
\end{figure}

% --- Para 2: existing objectives cannot see the split --------
Decision models aim to provide calibrated probabilities over possible answers~\citep{typesafe2026jev,PRDO}. Yet the same probabilities can arise from different patterns of judgment. Consider Van Gogh's \emph{Starry Night Over the Rh\^one}. Asked how a viewer feels, a model might assign equal probability to contentment and sadness. In the population at the top of Figure~\ref{fig:motivation}, half the viewers see a peaceful night by the river and feel content, while the other half are reminded of a distant home and feel sad. In the population at the bottom, every viewer is torn between the two emotions. The same prediction is calibrated in both cases. In the first, knowing how a viewer interprets the painting helps predict their response. In the second, the viewer may need more evidence to settle on a response. We therefore want a decision model to report which beliefs are present, how common they are, and how people holding each belief would answer.

Standard post-training objectives see the same training signal in the two populations in Figure~\ref{fig:motivation}. Supervised fine-tuning, reinforcement learning with verifiable rewards (RLVR), and PRDO all score a prediction made from the question alone against one person's observed choice at a time. Averaged over the population, the score depends only on how often each answer is chosen. In the painting example, contentment and sadness appear equally often in both populations, even though viewers arrive at those answers in different ways. The scoring rule determines what the model learns from these frequencies. An exact-match reward with GRPO favors the most frequent answer~\citep{shao2024deepseekmath}, while a proper scoring rule favors the pooled distribution. Group-aware methods use annotator labels to go beyond population-level frequencies. SubPOP conditions on demographic groups~\citep{suh2025subpop}, GDPO models stance distributions within groups~\citep{yao2025gdpo}, and P-GRPO normalizes advantages within user groups~\citep{wang2026pgrpo}. Yet people with the same label can interpret the same question differently~\citep{sap2022annotators,orlikowski2023ecological}. A prediction conditioned on that label still pools their choices, leaving open how each observed choice should inform the predictions for the beliefs that could have produced it.

An observed choice offers evidence about which belief may have produced it. We introduce \emph{Belief-Grouped Policy Optimization} (BGPO), a post-training method that uses this evidence to learn belief-specific choice distributions.
In the painting example, a response of ``sadness'' is more likely to come from a viewer reminded of home. 
BGPO uses Bayes' rule to estimate the probability that each belief generated an observed choice, incorporating annotator labels when available. 
These posterior probabilities weight the choice's contribution to each belief's log-score reward. 
BGPO also scores each belief list by how well its mixture predicts observed choices and annotator labels. To surface rare beliefs, it generates additional candidate lists from a small set of annotations and rewards them according to how well they predict held-out annotations.
At inference, the policy receives only the question and combines the belief-specific predictions according to their estimated prevalence.
We evaluate BGPO on five text and image benchmarks covering opinions, conversation safety, emotions elicited by artwork, and photo aesthetics. Across both model initializations, BGPO achieves the lowest conditional, worst-group, and overall Jensen--Shannon divergence, reducing conditional divergence by up to 44\% relative to the strongest baseline. To test whether BGPO also recovers the beliefs behind individual choices, we introduce CLEVR-Belief, a synthetic benchmark in which annotators follow known rules. On CLEVR-Belief, BGPO recovers the underlying belief groups more accurately than label-conditioned baselines, as measured by the adjusted Rand index.

Our contributions are threefold. First, we propose BGPO, a post-training method that uses posterior-weighted rewards to learn belief-specific choice distributions and held-out choices to evaluate candidate belief lists. Second, we introduce CLEVR-Belief and evaluation metrics for measuring whether a model recovers the underlying belief groups alongside the aggregate choice distribution. Third, we conduct a comparative evaluation across text and image tasks that assesses prediction quality at the conditional, worst-group, and population levels.

\section{Related Work}
\label{sec:related}

\paragraph{Post-training decision models.}
Decision models such as Jev output a probability for each option~\citep{typesafe2026jev}, and open recipes train them by supervised fine-tuning on observed choices~\citep{geuter2026releasing}. RLVR~\citep{lambert2024tulu3} instead rewards correct answers, typically with GRPO~\citep{shao2024deepseekmath} and its variants~\citep{yu2025dapo,liu2025drgrpo}, while calibration-oriented RL rewards a proper score of the stated confidence~\citep{baniharouni2025rewarding,damani2025beyond} or, in PRDO, of the answer distribution~\citep{PRDO}. RLHF and DPO~\citep{ouyang2022training,rafailov2023dpo} likewise fit one target per prompt and treat disagreement as noise~\citep{yao2025gdpo}, although it often reflects annotators' beliefs~\citep{sap2022annotators}. These objectives differ in their scoring rule, but each scores one prediction per question against one person's choice at a time and thus learns the pooled distribution, or only the majority answer. 
In contrast, BGPO predicts which beliefs are present, how common each is, and how each would choose, crediting every choice to the beliefs most likely to have produced it.

\paragraph{Pluralistic alignment.}
Pluralistic alignment asks a model to represent the range of views in a population~\citep{sorensen2024roadmap}, measured by how well its answer distribution matches a population or group~\citep{santurkar2023whose,meister2025benchmarking}. Most methods condition on the group: they prompt with a persona~\citep{argyle2022onemanyusinglanguage}, adapt from a group's examples~\citep{zhao2024gpo}, fine-tune on group-conditioned survey responses~\citep{suh2025subpop}, calibrate a group's distribution over predefined stances~\citep{yao2025gdpo}, reward a country's answer distribution with GRPO~\citep{pham2026evidencebased}, or optimize the worst-off group or normalize advantages per group~\citep{ramesh2024grpo,wang2026pgrpo}. A second line treats the source of disagreement as latent: DPL predicts a distribution of rewards~\citep{siththaranjan2024dpl}, MaxMin-RLHF and EM-DPO fit latent user types by EM and aggregate them into one policy~\citep{chakraborty2024maxmin,chidambaram2024emdpo}, and VPL and PAL infer a user's preference from that user's comparisons~\citep{poddar2024vpl}. OP-GRPO rewards one response for covering the perspectives people have written~\citep{fu2026overton}, without saying how common each is or how its holders choose. These methods need a group label or a user's history, or keep disagreement inside a reward model, and use it to personalize or aggregate rather than to describe the population. In contrast, BGPO trains the policy itself to report which beliefs are present, how common each is, and how each chooses, with no group at inference.

\paragraph{Disagreement and uncertainty.}
Work on human label variation argues that disagreement among annotators can reveal systematic differences in how people interpret and judge the same input~\citep{plank2022problem}. One way to preserve this variation is to train on the pooled label distribution~\citep{peterson2019human}. Another is to predict each annotator's response separately, which requires annotator identities~\citep{davani2021dealingdisagreementslookingmajority,gordon2022jurylearningintegratingdissenting}. In vision, the Probabilistic U-Net represents ambiguity by sampling multiple plausible segmentations of the same image from a latent code~\citep{DBLP:conf/nips/KohlRMFLMERR18}. Recent work also finds that vision--language models struggle to reproduce the spread of human answers~\citep{DBLP:conf/aaai/LanFP25}.

A separate line of work studies how much of a model's predictive uncertainty is reducible and how much is irreducible~\citep{DBLP:conf/nips/KendallG17,hüllermeier2020aleatoricepistemicuncertaintymachine}. In language models, this distinction has been studied by clarifying the input~\citep{DBLP:conf/icml/HouLQAC024} or through iterative prompting~\citep{DBLP:conf/nips/Abbasi-YadkoriK24}. These approaches characterize uncertainty in the model's prediction. BGPO asks how different beliefs in the population contribute to that prediction. It learns from observed choices without requiring annotator identities and reports which beliefs are present, how common they are, and how people holding each belief would answer.

\section{Method}
\label{sec:method}

\input{fig/method}

\subsection{Problem Setup}
\label{sec:setup}

A question $x$, given as text or as an image with text, has a set of options $\mathcal{Y}$. An annotation $(y,g)$ records one person's choice $y\in\mathcal{Y}$ and their label $g$ from a label set $\mathcal{G}$, such as demographic groups, and $\mathcal{D}_x$ is the set of annotations for $x$. We model the population behind $x$ as a mixture of finitely many latent beliefs $b\in\mathcal{B}_x$ with shares $P(b\mid x)$. Each person holds one belief, and their label and choice are drawn from it independently. The choice thus depends on the belief alone, and the label only tells which belief a person is likely to hold:
\begin{equation}
  P(y,g\mid x)=\textstyle\sum_{b}P(b\mid x)\,P(g\mid b,x)\,P(y\mid b,x).
  \label{eq:population}
\end{equation}
Existing post-training objectives output one distribution $p$ over $\mathcal{Y}$ per question and so miss this structure. Under a strictly proper score~\citep{gneiting2007strictly}, such as the log score of SFT and PRDO, the optimal $p$ is the pooled distribution $P(y\mid x)$; under a reward linear in $p$, such as that of RLVR, it is the most common choice.

Pooling costs exactly $I(Y;B\mid x)$ in average log score per annotation, the mutual information between a random person's choice $Y$ and belief $B$. Let $G$ be that person's label. Since the label says nothing about the choice beyond the belief, $I(Y;G\mid B,x)=0$, and the chain rule splits this cost into two parts:
\begin{equation}
  I(Y;B\mid x)=\underbrace{I(Y;G\mid x)}_{\text{between labels}}+\underbrace{I(Y;B\mid G,x)}_{\text{within labels}}.
  \label{eq:pooling}
\end{equation}
Conditioning on the label recovers only the first part. BGPO therefore predicts beliefs rather than labels, and uses labels only as evidence about them.

\subsection{Belief Lists and Answer Rewards}
\label{sec:rollout}

The policy $\pi_\theta$ describes the population of $x$ in two stages as shown in Figure \ref{fig:method}. First, it writes a list of beliefs $h\sim\pi_\theta(\cdot\mid x)$. We write $K(h)$ for the number of beliefs in $h$, which is at most a fixed cap $K_{\max}$. Each belief $k\in\{1,\dots,K(h)\}$ has a one-sentence description $d_k$, a share $w_k$, and a label distribution $\eta_k$ over $\mathcal{G}$ that says who tends to hold it; the prompt lists $\mathcal{G}$ but gives no annotator's label. Second, the policy answers $n$ times as a holder of each belief, $z_{kj}\sim\pi_\theta(\cdot\mid x,h,k)$. Each answer gives a short reasoning and a choice distribution $u_z$ over $\mathcal{Y}$. The description $d_k$ is not scored itself, but it shapes the answers for belief $k$.

Averaging the $n$ answers gives $\bar U_k=\frac1n\sum_{j=1}^{n}u_{z_{kj}}$, an estimate of belief $k$'s prediction $U_k=\mathbb{E}_{z\sim\pi_\theta(\cdot\mid x,h,k)}[u_z]$. Mirroring \cref{eq:population}, a list with its answers, which we call a rollout, predicts
\begin{equation}
  \hat P_h(y,g)=\textstyle\sum_{k=1}^{K(h)}w_k\,\eta_k(g)\,\bar U_k(y).
  \label{eq:mixture}
\end{equation}
Its marginal $\hat P_h(y)$ is the label-free prediction used at deployment, $\hat P_h(g)$ is the composition of the population, and $\hat P_h(y\mid g)$ is the prediction for label $g$.

Each training step splits $\mathcal{D}_x$ into two disjoint sets of $M$ annotations. Rollouts are scored on $\Deval$, and we write $\hat{\mathbb{E}}_{\Deval}$ for the mean over $(y,g)\in\Deval$. $\Dobs$ is read only by the posterior over lists introduced below, so no list is rewarded for copying annotations it has read.

\paragraph{Crediting each choice.}
\phantomsection\label{sec:credit}
Training must decide which annotations each belief learns from. Suppose belief $k$ learns from the $m$-th annotation $(y_m,g_m)$ of $\Deval$ with weight $\omega_k^m\ge0$. Maximizing $\sum_m\omega_k^m\log U_k(y_m)$ gives $U_k^\star(y)\propto\sum_m\omega_k^m\mathbf{1}[y_m=y]$, the weighted frequency of each choice. A single output uses $\omega_k^m=1$ and learns the pooled distribution. Label conditioning keeps one prediction per label, so $k$ ranges over $\mathcal{G}$; it uses $\omega_k^m=\mathbf{1}[k=g_m]$ and learns the pooled distribution within each label. Neither weight looks at the choice $y_m$. BGPO uses Bayes' rule instead: the responsibility of belief $k$ for $(y,g)$ is the probability, under \cref{eq:mixture}, that $k$ produced it,
\begin{equation}
  r_k(y,g)=w_k\,\eta_k(g)\,\bar U_k(y)\,/\,\hat P_h(y,g).
  \label{eq:resp}
\end{equation}
A choice is credited mostly to the beliefs that predict it and are common under its label. With $\omega_k^m=r_k(y_m,g_m)$, $U_k^\star$ is the EM update for the mixture~\citep{dempster1977em}, and the gradient of its log-likelihood weights belief $k$ by the same $r_k$. Holding $r_k$ fixed as in EM, each answer is rewarded by its log score on the annotations credited to its belief:
\begin{equation}
  S(z_{kj})=\hat{\mathbb{E}}_{\Deval}\bigl[r_k(y,g)\,\log u_{z_{kj}}(y)\bigr].
  \label{eq:answer-reward}
\end{equation}

\paragraph{Grouping answers by belief.}
The $n$ answers for one belief of one list share the prompt $(x,h,k)$ and the weights $r_k$, so we compare them only with each other. Each gets the leave-one-out advantage $A(z_{kj})=S(z_{kj})-\frac{1}{n-1}\sum_{j'\neq j}S(z_{kj'})$~\citep{ahmadian2024back}. An unparsable answer is left out of $\bar U_k$ and scores lowest in its group. Pooling answers across beliefs would fail. As $\log u\le0$, an answer's score is closer to zero the less total credit $\hat{\mathbb{E}}_{\Deval}[r_k]$ its belief receives, so answers for rare beliefs would win regardless of quality. Grouping by belief removes this offset and gives the method its name. With one belief and no posterior, $r\equiv1$ and $S$ reduces to the log-score reward of PRDO~\citep{PRDO}, applied here to a stated rather than a read-out distribution.

\subsection{List Rewards and Posterior Lists}
\label{sec:posterior}

A list and its answers are rewarded by how well they predict $\Deval$:
\begin{equation}
  R(h)=\hat{\mathbb{E}}_{\Deval}\bigl[\log\hat P_h(y,g)+\alpha\log\hat P_h(y)\bigr]-\lambda K(h).
  \label{eq:list-reward}
\end{equation}
The joint log score is strictly proper~\citep{gneiting2007strictly}: its expectation is maximized exactly when $\hat P_h(y,g)=P(y,g\mid x)$, so it rewards shares, label distributions and answers together. The term weighted by $\alpha\ge0$ stresses the label-free prediction used at deployment, so an imperfect list does not trade it for a better label fit. The joint log score alone does not single out one list: every exact factorization of $P(y,g\mid x)$ attains the maximum, including the label-conditional list with one belief per label, $\eta_k(g)=\mathbf{1}[g=k]$. The cost $\lambda\ge0$ per belief prefers the list with the fewest beliefs, which rules out the label-conditional list whenever fewer beliefs than labels explain the annotations. Remaining ties are broken by the policy's prior over lists, and we measure on synthetic data whether the true beliefs are recovered. Lists that are unparsable, truncated or longer than $K_{\max}$ get a constant reward below that of any parsed list.

% float placed early so that it lands next to the results
\begin{table*}[t]
  \centering
  \scriptsize
\caption{\textbf{Results on text benchmarks.}
Cond., Worst, and All report mean per-label, worst-group, and label-free JSD, respectively. Rec.\ denotes recovery rate.
Label indicates use of annotator labels during training. All methods perform inference without annotator labels. Results are averaged over three seeds.}
  \label{tab:main-text}
  \resizebox{\textwidth}{!}{%
  \begin{tabular}{@{}lc cccc cccc cccc@{}}
    \toprule
    & & \multicolumn{4}{c}{OpinionQA} & \multicolumn{4}{c}{GlobalOpinionQA} & \multicolumn{4}{c}{DICES} \\
    \cmidrule(lr){3-6}\cmidrule(lr){7-10}\cmidrule(lr){11-14}
    Method & Label & Cond.\,$\downarrow$ & Worst\,$\downarrow$ & All\,$\downarrow$ & Rec.\,$\uparrow$ & Cond.\,$\downarrow$ & Worst\,$\downarrow$ & All\,$\downarrow$ & Rec.\,$\uparrow$ & Cond.\,$\downarrow$ & Worst\,$\downarrow$ & All\,$\downarrow$ & Rec.\,$\uparrow$ \\
    \midrule
    \multicolumn{14}{@{}l}{\textit{Released decision model}} \\
    Jev (zero-shot)~\citep{typesafe2026jev} & -- & .0854 & .1394 & .1179 & -- & .1311 & .2040 & .1725 & -- & .1032 & .2023 & .0958 & -- \\
    \midrule
    \multicolumn{14}{@{}l}{\textbf{Initialized from Qwen3.5-2B}} \\
    Zero-shot & -- & .0755 & .0815 & .0661 & -- & .0968 & .1065 & .0799 & -- & .0867 & .1322 & .0462 & -- \\
    \addlinespace
    SFT~\citep{geuter2026releasing} & $\times$ & .0571 & .0655 & .0484 & -- & .0841 & .0984 & .0676 & -- & .0731 & .1490 & .0280 & -- \\
    
    PRDO~\citep{PRDO} & $\times$ & .0548 & .0640 & .0459 & -- & .0797 & .0905 & .0629 & -- & .0793 & .1623 & .0326 & -- \\
    \addlinespace
    GRPO~\citep{shao2024deepseekmath} & $\times$ & .1121 & .1206 & .1039 & -- & .1386 & .1507 & .1239 & -- & .1103 & .1830 & .0777 & -- \\
    GRPO + label & \checkmark & .1352 & .1396 & .1160 & -3.5115 & .1321 & .1419 & .1021 & -2.0425 & .1186 & .2094 & .0548 & -1.0288 \\
    \addlinespace
    Label-cond.\ SFT & \checkmark & .0617 & .0710 & .0529 & .0274 & .0835 & .0981 & .0672 & .0204 & .0698 & .1434 & .0270 & -.0141 \\
    
    Two-stage SFT & \checkmark & .0617 & .0710 & .0529 & .0263 & .0835 & .0981 & .0672 & .0190 & .0698 & .1434 & .0272 & -.0067 \\
    
    PRDO + label & \checkmark & .0560 & .0607 & .0474 & .0460 & .0735 & .0850 & .0570 & .0102 & .0814 & .1697 & .0316 & -.0475 \\
    
    GDPO~\citep{yao2025gdpo} & \checkmark & .0569 & .0644 & .0477 & .0134 & .0780 & .0892 & .0613 & -.0104 & .0741 & .1301 & .0336 & -.0204 \\
    \addlinespace
    \rowcolor{gray!10}\textbf{BGPO (ours)} & \checkmark & .0350 & .0419 & .0267 & .1257 & .0516 & .0590 & .0360 & .0274 & .0422 & .0775 & .0127 & .2986 \\
    \midrule
    \multicolumn{14}{@{}l}{\textbf{Initialized from Qwen3.5-4B}} \\
    Zero-shot & -- & .0699 & .0815 & .0599 & -- & .1036 & .1165 & .0874 & -- & .0729 & .1000 & .0384 & -- \\
    \addlinespace
    SFT~\citep{geuter2026releasing} & $\times$ & .0450 & .0526 & .0357 & -- & .0626 & .0761 & .0462 & -- & .0560 & .0979 & .0209 & -- \\
    
    PRDO~\citep{PRDO} & $\times$ & .0454 & .0551 & .0361 & -- & .0651 & .0793 & .0489 & -- & .0678 & .1397 & .0283 & -- \\
    \addlinespace
    GRPO~\citep{shao2024deepseekmath} & $\times$ & .1904 & .2073 & .1824 & -- & .2350 & .2574 & .2248 & -- & .1441 & .1609 & .1285 & -- \\
    
    GRPO + label & \checkmark & .1591 & .1671 & .0955 & -17.7649 & .1630 & .1831 & .1057 & -8.3106 & .1001 & .1264 & .0664 & -2.9455 \\
    \addlinespace
    Label-cond.\ SFT & \checkmark & .0454 & .0490 & .0376 & .1661 & .0576 & .0673 & .0423 & .0743 & .0671 & .1465 & .0215 & -.0206 \\
    
    Two-stage SFT & \checkmark & .0454 & .0490 & .0378 & .1875 & .0576 & .0673 & .0423 & .0779 & .0671 & .1465 & .0221 & .0021 \\
    
    PRDO + label & \checkmark & .0441 & .0476 & .0369 & .2207 & .0664 & .0864 & .0515 & .0995 & .0630 & .1311 & .0211 & -.0151 \\
    
    GDPO~\citep{yao2025gdpo} & \checkmark & .0428 & .0460 & .0352 & .2034 & .0651 & .0793 & .0491 & .0431 & .0655 & .1379 & .0241 & .0046 \\
    \addlinespace
    \rowcolor{gray!10}\textbf{BGPO (ours)} & \checkmark & .0240 & .0283 & .0196 & .5581 & .0447 & .0523 & .0306 & .1288 & .0402 & .0776 & .0101 & .2763 \\
    \bottomrule
  \end{tabular}}
\end{table*}

\paragraph{Posterior over lists.}
On-policy sampling rarely finds a small belief. Let $E_b$ be the set of lists naming a belief $b$ of share $\rho=P(b\mid x)$. Then the $L$ lists sampled from the policy for $x$ include one from $E_b$ with probability at most $L\,\pi_\theta(E_b\mid x)$, which is small if the policy seldom writes $b$. Yet $\Dobs$ already contains a holder of $b$ with probability $1-(1-\rho)^M$. We therefore let the policy also propose $H$ lists per question after reading $\Dobs$ as (choice, label) pairs, much as STaR~\citep{zelikman2022star} and TRICE~\citep{phan2023trice} propose rationales after seeing the answer. The same network serves as this posterior $q_\theta(\hat h\mid x,\Dobs)$, so no parameters are added, as in LaTRO~\citep{chen2024latro}. A posterior list is rewarded by how well it explains $\Dobs$, minus a penalty of weight $\beta>0$ for moving away from the policy:
\begin{equation}
  R_q(\hat h)=\log\hat P_{\hat h}(\Dobs)-\beta\log\frac{q_\theta(\hat h\mid x,\Dobs)}{\pi_\theta(\hat h\mid x)},
  \label{eq:post-reward}
\end{equation}
where $\hat P_{\hat h}(\Dobs)=\prod_{(y,g)\in\Dobs}\hat P_{\hat h}(y,g)$, and both sequence log-probabilities come from teacher forcing and are held constant. In expectation the penalty is a KL term, and the optimal posterior is $q^\star(h)\propto\pi_\theta(h\mid x)\,e^{f(h)/\beta}$ with $f(h)=\mathbb{E}_z[\log\hat P_h(\Dobs)]$, the expected log-likelihood over the answers $z$ sampled for $h$~\citep{korbak2022rl}. At $\beta=1$, $q^\star$ is the Bayes posterior over lists with the policy as prior; our $\beta<1$ tempers it, so the annotations count $1/\beta$ times. The penalty keeps proposals close to the policy, which can absorb them only gradually: the clipped objective in \cref{eq:surrogate} limits each token's probability ratio to $1+\epsilon_{\text{high}}$ per update~\citep{yu2025dapo}. The $H$ posterior lists for one $(x,\Dobs)$ get leave-one-out advantages $A_q(\hat h)$ from $R_q$.

\paragraph{Mixed-policy groups and optimization.}
As in LUFFY's mixed-policy GRPO~\citep{yan2025luffy}, the $L$ policy lists and $H$ posterior lists of a question form one group, and each list gets the leave-one-out advantage $A(h)$ of its reward $R(h)$. A posterior list thus gains probability under the policy's prompt $x$ only if it beats the group on $\Deval$. 
We omit the importance weight $\pi_\theta(\hat h\mid x)/q_\theta(\hat h\mid x,\Dobs)$, since it is smallest for lists describing beliefs the policy lacks, which are exactly the lists the posterior exists to supply. 
The old probability $\pi_{\thetaold}(\hat h\mid x)$ of a posterior list, with $\thetaold$ the sampling parameters, comes from teacher forcing, so both kinds of list share one clipped ratio. All lists are answered from $(x,h,k)$, which contains no annotations, so each list is judged by answers the deployed policy could give.

One clipped surrogate~\citep{schulman2017ppo,shao2024deepseekmath} trains three parts together. Each sampled sequence $o$ enters the batch $\mathcal{O}$ with the context $c$ it is trained under and its advantage $A$: every list as $(h,x,A(h))$, every posterior list also as $(\hat h,(x,\Dobs),A_q(\hat h))$, and every answer as $(z_{kj},(x,h,k),A(z_{kj}))$. With the token ratio $s_t=\pi_\theta(o_t\mid c,o_{<t})/\pi_{\thetaold}(o_t\mid c,o_{<t})$ and its clipped value $\bar s_t=\clip(s_t,1-\epsilon_{\text{low}},1+\epsilon_{\text{high}})$, we maximize
\begin{equation}
  \mathcal{J}(\theta)=\frac{1}{N_{\mathcal{O}}}\sum_{(o,c,A)\in\mathcal{O}}\;\sum_{t=1}^{|o|}\min\bigl(s_tA,\;\bar s_tA\bigr),
  \label{eq:surrogate}
\end{equation}
where $N_{\mathcal{O}}$ is the number of tokens in the batch. The mean over tokens and the wider upper clip, $\epsilon_{\text{high}}>\epsilon_{\text{low}}$, follow DAPO~\citep{yu2025dapo}. As in Dr.~GRPO~\citep{liu2025drgrpo}, we do not divide advantages by the group's standard deviation, which would give every belief's answers updates of one size and undo the responsibility weighting. As in DAPO, we add no KL toward the base model, which states one pooled distribution and would pull lists back toward a single belief.

% float placed early so that it lands next to the results

\begin{table*}[t]
  \centering
  \small
  \setlength{\tabcolsep}{2.5pt}
 \caption{\textbf{Results on ArtELingo and PARA.}
Cond., Worst, and All report mean per-label, worst-group, and label-free JSD, respectively. Rec.\ denotes recovery rate.
Label indicates whether training uses annotator labels. All methods perform inference without annotator labels.
We report zero-shot Jev with image and text-only inputs; its rows repeat the same numbers under both initial models.}
  \label{tab:main-image}
  \resizebox{\textwidth}{!}{%
    \begin{tabular}{@{}lc cccc cccc @{\hspace{10pt}} cccc cccc@{}}
    \toprule
    & & \multicolumn{8}{c}{Qwen3.5-2B} & \multicolumn{8}{c}{Qwen3.5-4B} \\
    \cmidrule(lr){3-10}\cmidrule(lr){11-18}
    & & \multicolumn{4}{c}{ArtELingo} & \multicolumn{4}{c}{PARA} & \multicolumn{4}{c}{ArtELingo} & \multicolumn{4}{c}{PARA} \\
    \cmidrule(lr){3-6}\cmidrule(lr){7-10}\cmidrule(lr){11-14}\cmidrule(lr){15-18}
    Method & Label & Cond.\,$\downarrow$ & Worst\,$\downarrow$ & All\,$\downarrow$ & Rec.\,$\uparrow$ & Cond.\,$\downarrow$ & Worst\,$\downarrow$ & All\,$\downarrow$ & Rec.\,$\uparrow$ & Cond.\,$\downarrow$ & Worst\,$\downarrow$ & All\,$\downarrow$ & Rec.\,$\uparrow$ & Cond.\,$\downarrow$ & Worst\,$\downarrow$ & All\,$\downarrow$ & Rec.\,$\uparrow$ \\
    \midrule
    \multicolumn{18}{@{}l}{\textit{Released decision model} (the same model in both blocks)} \\
    Jev (zero-shot)~\citep{typesafe2026jev} & -- & .4042 & .4352 & .3088 & -- & .3393 & .6989 & .2300 & -- & .4042 & .4352 & .3088 & -- & .3393 & .6989 & .2300 & -- \\
    Jev (text only) & -- & .8412 & .9150 & .8142 & -- & .5547 & .8596 & .5362 & -- & .8412 & .9150 & .8142 & -- & .5547 & .8596 & .5362 & -- \\
    \midrule
    Zero-shot & -- & .6774 & .7607 & .5807 & -- & .7467 & .8890 & .6820 & -- & .5724 & .6163 & .4544 & -- & .6284 & .8685 & .5408 & -- \\
    \addlinespace
    SFT~\citep{geuter2026releasing} & $\times$ & .3342 & .3531 & .1606 & -- & .2464 & .6331 & .1084 & -- & .3160 & .3677 & .1562 & -- & .2213 & .5671 & .0862 & -- \\
    PRDO~\citep{PRDO} & $\times$ & .3246 & .3546 & .1604 & -- & .2465 & .5805 & .1129 & -- & .3198 & .3624 & .1534 & -- & .2375 & .5706 & .1040 & -- \\
    \addlinespace
    GRPO~\citep{shao2024deepseekmath} & $\times$ & .4443 & .4583 & .3439 & -- & .3774 & .8202 & .2626 & -- & .4530 & .5052 & .3794 & -- & .4035 & .7136 & .2956 & -- \\
    GRPO + label & \checkmark & .4282 & .4583 & .2959 & -1.0334 & .4761 & .7431 & .3299 & -.7668 & .4632 & .4860 & .3465 & -1.4613 & .4287 & .7938 & .3074 & -.8895 \\
    \addlinespace
    Label-cond.\ SFT & \checkmark & .3305 & .3721 & .1728 & .0151 & .2405 & .6342 & .1060 & -.0070 & .3348 & .3691 & .1646 & .0113 & .2196 & .5636 & .0849 & -.0029 \\
    Two-stage SFT & \checkmark & .3302 & .3721 & .1725 & .0154 & .2408 & .6330 & .1065 & -.0055 & .3350 & .3693 & .1648 & .0113 & .2196 & .5636 & .0850 & -.0013 \\
    PRDO + label & \checkmark & .3200 & .3465 & .1550 & .0007 & .2353 & .5585 & .1007 & -.0123 & .3444 & .3667 & .1686 & .0056 & .2321 & .5253 & .1032 & .0108 \\
    GDPO~\citep{yao2025gdpo} & \checkmark & .3231 & .3720 & .1625 & .0212 & .2345 & .5816 & .0979 & -.0096 & .3180 & .3616 & .1522 & .0072 & .2220 & .5596 & .0869 & -.0033 \\ 
    \addlinespace
    \rowcolor{gray!10}\textbf{BGPO (ours)} & \checkmark & .3043 & .3352 & .1398 & .0621 & .2061 & .4862 & .0773 & .0264 & .3020 & .3287 & .1401 & .0716 & .2034 & .4784 & .0754 & .0258 \\
    \bottomrule
  \end{tabular}}
\end{table*}

\section{Experiment}
\label{sec:exp}

\subsection{Experiment Setup}
\label{sec:exp-setup}

\paragraph{Datasets.}
We evaluate on text and image benchmarks in which multiple annotators answer the same question and each response comes with an annotator label. For text, OpinionQA~\citep{santurkar2023whose} contains 1,498 multiple-choice questions from Pew surveys, with responses labeled by political party. GlobalOpinionQA~\citep{durmus2024towards} contains 2,556 survey questions, with responses labeled by country. DICES~\citep{aroyo2023dices} contains chatbot conversations rated for safety by annotators from diverse demographic groups. For images, ArtELingo~\citep{mohamed2022artelingo} asks annotators to select one of nine emotions evoked by a WikiArt painting and uses annotation language as a proxy for culture. PARA~\citep{yang2022para} contains 31,220 photos rated for aesthetics on a 1--5 scale by 438 annotators, whose labels indicate their photographic experience.

We split all datasets by question and retain real-world benchmarks whose annotator labels are informative about choices according to a permutation test. The supplement provides dataset statistics, preprocessing details, and the selection procedure. Because the beliefs behind individual choices are unobserved in these benchmarks, we introduce CLEVR-Belief, a synthetic benchmark built on CLEVR scenes~\citep{johnson2017clevr} that lets us evaluate belief recovery directly.

\paragraph{CLEVR-Belief.}
Each question presents a CLEVR scene with three or four marked objects and asks which object a rater would pick first. Each synthetic rater follows one of four rules. The rater picks the large, metal, red, or leftmost object. We treat each rule as a distinct belief. Two parameters control the difficulty of recovering these beliefs. We set the prevalence of the rarest rule to $\rho_{\min}=0.1$ to test whether the model can recover a belief held by a small minority. We set $\kappa=0.7$ so that each rater's label matches their rule with probability $\kappa$. Each label therefore mixes raters who follow different rules, allowing us to test whether the model can recover the underlying beliefs from choices and imperfect labels. The supplement describes the construction procedure.

\paragraph{Baselines.}
All methods start from the same initial model and use the same number of training steps, the same tuning budget and the same number of annotations per question and step, and all RL methods share one trainer.
For each initial model, its zero-shot prediction is a reference row, and the released decision model Jev~\citep{typesafe2026jev} is an external reference. We group the trained baselines by what they learn. \emph{Decision-model recipes} train the initial model to state one distribution over options per question: SFT on the empirical choice distribution, the open recipe for decision models~\citep{geuter2026releasing}, and PRDO~\citep{PRDO}, which rewards the log score of that distribution. \emph{RLVR} is GRPO with a 0/1 reward~\citep{shao2024deepseekmath}, on images following Visual-RFT and VLM-R1~\citep{liu2025visualrft,shen2025vlmr1}, with and without the label in the prompt. \emph{Label-conditioned methods} train with the annotator label: label-conditional SFT, a two-stage model that adds a predictor of the label shares, PRDO with the label in the prompt, and GDPO~\citep{yao2025gdpo}.
No method receives a label at inference. A method trained with labels therefore predicts the share of each label and averages its per-label predictions with these shares.

%with the label, which sees the same annotations as BGPO, is the closest comparison.

% Figure 3 placed here so that it lands at the top of the next page
\input{fig/cases}

\paragraph{Metrics.}
Every metric compares a predicted distribution over the options with the choices annotators made on test questions. The first three use the Jensen--Shannon divergence (JSD), which is 0 when the prediction matches the human choices exactly and grows with the mismatch. \emph{Overall} JSD compares the label-free prediction with the choices of all annotators pooled, and measures how well a method predicts the population as a whole. \emph{Conditional} JSD compares the prediction for each label with the choices of the annotators carrying that label, averaged uniformly over labels, and measures how well a method predicts each group. \emph{Worst-group} JSD is the largest of these per-label values, and shows whether a low average hides a poorly predicted group. The \emph{recovery rate} (Rec.) measures how much of the label's information a method uses. It is the gain in log score when a method predicts with the label rather than without it, divided by the same gain computed from the human choices themselves, which is the largest gain the label allows. The supplement gives its formal definition and explains how to read the values of each metric.

On synthetic data each rater's belief is known, so two further metrics score the beliefs themselves. \emph{Small.} is the conditional JSD of the rarest belief after one-to-one matching of predicted and true beliefs. \emph{ARI} is the adjusted Rand index~\citep{hubert1985comparing} between the belief with the highest responsibility for each annotation and the rater's true belief.

\paragraph{Implementation details.}
We initialize all trained methods from Qwen3.5-2B and Qwen3.5-4B~\citep{qwen2026qwen35}, both of which support text and image inputs. We report results for both model sizes in \cref{tab:main-text,tab:main-image} and use Qwen3.5-4B for CLEVR-Belief (\cref{tab:clevr}) and the ablation study in the supplement. We train all RL methods with verl~\citep{sheng2025hybridflow}.
At each training step, BGPO generates $L=4$ policy lists and $H=4$ posterior lists per question. Each list contains up to $K_{\max}=4$ beliefs, with $n=2$ sampled answers per belief. The observed and held-out sets, $\Dobs$ and $\Deval$, each contain $M=4$ annotations. We set $\alpha=\beta=0.5$ and use asymmetric clipping with $(\epsilon_{\text{low}},\epsilon_{\text{high}})=(0.2,0.28)$. To keep log scores finite, we floor predicted probabilities at $\delta=0.01$ before renormalizing. We apply the per-belief penalty ($\lambda=0.02$) starting at step 100, giving the policy time to learn predictions for multiple beliefs before penalizing longer lists.
Per question per training step, BGPO generates up to approximately 13.6k tokens, compared with approximately 4.1k for each baseline using a group of eight responses. We average results over three random seeds and provide the remaining training settings in the supplement.

\subsection{Main Results}
\label{sec:main-result}

\paragraph{Text benchmarks.}
As shown in \cref{tab:main-text}, BGPO achieves the lowest conditional, worst-group, and overall JSD across all three text benchmarks and both model initializations.
It reduces conditional JSD by up to 44\% relative to the strongest baseline in each setting. Conditioning on annotator labels yields little improvement in JSD for the baselines. BGPO recovers substantially more of the information carried by those labels on OpinionQA and DICES, with smaller gains on GlobalOpinionQA. GRPO is the only trained method with higher conditional JSD than the zero-shot model. Its exact-match reward concentrates probability on the most frequent answer, and its nearly one-hot label-conditioned predictions score worse than their mixture, yielding a negative recovery rate. On OpinionQA, initializing BGPO with the larger model further reduces conditional JSD.

\paragraph{Image benchmarks.}
BGPO achieves the lowest conditional, worst-group, and overall JSD on both image benchmarks and with both model initializations (\cref{tab:main-image}). The gains are smaller than on the text benchmarks, with conditional JSD falling by up to 12\% relative to the strongest baseline. BGPO improves worst-group performance most on PARA and achieves higher recovery rates than the baselines on both datasets, though the rates remain modest. On ArtELingo, conditional JSD is substantially higher than overall JSD across trained methods. Each painting has few annotations per language, making the per-language reference distributions noisy. Jev's conditional JSD rises sharply when the image is withheld, showing that the visual input contributes to its predictions.

\paragraph{Known beliefs.}
On the real-world benchmarks, JSD measures how closely BGPO's predicted response distributions match observed choices across the population and within each annotator label. Different belief lists can yield the same predicted distributions and thus the same JSD, even when they attribute responses to different beliefs. In CLEVR-Belief, each synthetic rater follows a known rule, and we control how closely annotator labels correspond to those rules. This setup lets us measure how accurately BGPO recovers the underlying groups and predicts the choices of raters in the rarest group.

\begin{wraptable}{r}{0.54\linewidth}
  \vspace{-10pt}
  \centering
  \footnotesize
  \setlength{\tabcolsep}{4pt}
\caption{\textbf{Results on CLEVR-Belief.}
Each label mixes raters of different rules ($\kappa=0.7$).
Small.\ reports conditional JSD for the smallest belief after matching.
ARI measures agreement with the true rules.}
  \label{tab:clevr}
    \begin{tabular}{@{}l ccc@{}}
    \toprule
    Method & Cond.\,$\downarrow$ & Small.\,$\downarrow$ & ARI\,$\uparrow$ \\
    \midrule
    \multicolumn{4}{@{}l}{\textit{Released decision model}} \\
    Jev (zero-shot)~\citep{typesafe2026jev} & 0.2705 & 0.5986 & 0.3730 \\
    \midrule
    Label-conditional SFT & 0.0642 & 0.3612 & 0.3730 \\
    PRDO + label & 0.1044 & 0.4993 & 0.3730 \\
    \midrule
    BGPO w/o posterior & 0.0841 & 0.1163 & 0.9865 \\
    \textbf{BGPO (ours)} & \textbf{0.0369} & \textbf{0.0176} & \textbf{0.9982} \\
    \bottomrule
  \end{tabular}
  \vspace{-6pt}
\end{wraptable}
As shown in \cref{tab:clevr}, BGPO recovers the underlying rater groups more accurately than label-conditioned baselines and better predicts the choices of raters who follow the rarest rule. Each annotator label mixes raters who follow different rules, so assigning annotations by label yields a low adjusted Rand index (ARI). BGPO achieves near-perfect agreement with the true rule assignments while also reducing conditional JSD.
% Removing posterior lists leaves the ARI largely unchanged, but substantially increases both conditional JSD and the divergence for the rarest rule.

\paragraph{Effect of posterior lists.}
ARI measures whether annotations are assigned to the correct belief groups, whereas Small.\ measures how well the model predicts choices within the rarest group. In \cref{tab:clevr}, removing posterior lists has little effect on ARI but substantially increases divergence for the rarest rule. This suggests that lists proposed from the question alone already separate the rater groups on CLEVR-Belief. Proposing additional lists from observed choices during training helps the model predict how raters who follow the rarest rule will answer.

\paragraph{Effect of responsibility credit.}
We also ablate responsibility credit in the supplement, crediting each annotation by its label alone instead of by Bayes' rule. On CLEVR-Belief, this more than doubles conditional JSD and reduces the recovery rate to zero, a larger loss than removing posterior lists. Crediting choices by responsibility is therefore the component that lets BGPO separate the beliefs mixed inside a label. Without it, the label adds nothing to the model’s predictions, which is what a recovery rate of zero means.

\subsection{Case Study}
\label{sec:cases}

We show the belief lists that BGPO writes for three ArtELingo test paintings in \cref{fig:cases}. The beliefs describe different readings of the same image, and each reading leads to a different emotion. For the landscape, viewers who wonder about the history of the ruined city pick awe and are most often Arabic-language annotators, while viewers who see a peaceful view pick contentment and are most often Chinese-language annotators. The annotations show the same split: 80\% of the Arabic-language annotators choose awe and all Chinese-language annotators choose contentment. For the dog portrait, the belief about getting closer to pets picks contentment and the belief about childhood memories picks amusement, and contentment is predicted at 56\% for Chinese-language annotators against 50\% observed.
For the engraving, the largest belief reads the scene as terrifying and picks fear, the most common choice of both the Chinese-language 80\% and the English-language 40\% annotators. 
The landscape example also shows two mismatches. BGPO predicts that 41\% of Arabic-language annotators would choose awe, compared with 80\% in the observed annotations. It also assigns a prevalence of 37\% to a belief that favors sadness, an emotion no annotator selected.

\section{Conclusion}

Existing post-training methods for decision models learn from pooled annotator choices, which can obscure whether a prediction reflects disagreement across people or uncertainty within individuals. We introduced Belief-Grouped Policy Optimization (BGPO), which learns to describe the beliefs present in a population, estimate how common they are, and predict how people holding each belief would respond. BGPO uses posterior proposals to surface beliefs that may be missed from the question alone and assigns credit for observed responses to the beliefs most likely to explain them. Across text and image benchmarks, BGPO improves both aggregate and group-level predictions and reduces error for the worst-predicted group, and recovers known beliefs on a synthetic task, while requiring only the question at inference. By making the source of uncertainty explicit, BGPO can help distinguish when to ask which interpretation applies and when to gather more evidence. 
Future work will extend BGPO to open-ended answers, where beliefs cannot be scored against a fixed option set.
\section*{LLM Usage Statement}
We used Claude Code for language editing, grammar
correction, and LaTeX formatting of the manuscript,
as well as for drafting code snippets and debugging scripts.

%% file: fig/method.tex
% Method figure: BGPO on one question, read left to right in four steps.
% One policy (robot + flame = the only trained model) writes belief lists, answers as each
% belief, is scored on annotations, and is updated from advantages computed within groups.
\definecolor{bgedge}{RGB}{64,64,64}
\definecolor{bgpolicy}{RGB}{251,229,214}
\definecolor{bgpanel}{RGB}{243,244,246}
\definecolor{bgflow}{RGB}{147,197,253}
\definecolor{bgfire}{RGB}{197,90,17}
\definecolor{bA}{RGB}{108,142,191}
\definecolor{bB}{RGB}{215,155,0}
\definecolor{bC}{RGB}{130,179,102}
\definecolor{bD}{RGB}{150,115,166}
\definecolor{fA}{RGB}{218,232,252}
\definecolor{fC}{RGB}{213,232,212}
\definecolor{fD}{RGB}{225,213,231}
\definecolor{fY}{RGB}{255,242,204}
\definecolor{yB}{RGB}{214,182,86}
\definecolor{bgup}{RGB}{130,179,102}
\definecolor{bgdown}{RGB}{184,84,80}
\tikzset{
  >=Stealth,
  every node/.style={font=\small},
  inbox/.style={draw=bgedge,fill=white,rounded corners=3pt,line width=0.6pt,align=center,
                minimum width=22mm,inner sep=4pt},
  policy/.style={draw=bgfire!70,fill=bgpolicy,rounded corners=5pt,line width=0.9pt,align=center,
                 minimum width=23mm,minimum height=30mm,font=\normalsize\bfseries},
  chip/.style={draw=bgedge,fill=white,rounded corners=3pt,line width=0.6pt,
               minimum width=37mm,minimum height=9.5mm,inner sep=0pt},
  arr/.style={->,line width=0.6pt,draw=bgedge},
  flow/.style={line width=1.3pt,draw=black!75,-{Stealth[length=3mm,width=2.4mm]}},
  ttl/.style={font=\normalsize,text=black,anchor=west,inner sep=0pt,text height=1.6ex,text depth=0.5ex},
  num/.style={circle,draw=black!75,fill=white,text=black,font=\small,line width=0.5pt,inner sep=1pt,minimum size=4.6mm},
}
% one belief of a list: who holds it, its description (grey lines), its share
\newcommand{\brow}[5]{% y, colour, bar length, share, description
  \node[text=#2,font=\normalsize] at (\bx+0.32,#1) {\faUserCircle};
  \node[font=\scriptsize,anchor=west,inner sep=0pt] at (\bx+0.62,#1) {#5};
  \fill[#2!20,rounded corners=1pt] (\bx+3.92,#1-0.11) rectangle ++(0.7,0.22);
  \fill[#2,rounded corners=1pt] (\bx+3.92,#1-0.11) rectangle ++(#3*0.875,0.22);
  \node[font=\scriptsize,anchor=west,inner sep=1pt] at (\bx+4.66,#1) {#4};
}
% a stack of list cards
\newcommand{\cardstack}[3]{% name, y top, y bottom
  \foreach \o in {0.18,0.09}
    \draw[draw=bgedge!45,fill=white,rounded corners=3pt,line width=0.5pt]
      (\bx+\o,#2+\o) rectangle (\bx+5.3+\o,#3+\o);
  \draw[draw=bgedge,fill=white,rounded corners=3pt,line width=0.6pt] (\bx,#2) rectangle (\bx+5.3,#3);
  \coordinate (#1w) at (\bx,{(#2+#3)/2});
}
% the four options as faces: contentment, sadness, awe, fear
\definecolor{emo}{RGB}{255,205,60}
\newcommand{\emo}[1]{{\rlap{\color{emo}\faCircle}\color{black!80}#1}}
\newcommand{\fa}{\emo{\faSmile[regular]}}\newcommand{\fb}{\emo{\faSadTear[regular]}}\newcommand{\fc}{\emo{\faSurprise[regular]}}\newcommand{\fd}{\emo{\faGrimace[regular]}}
% one answer: a distribution over the four options
\newcommand{\ans}[7]{% x, y, colour, four bar heights
  \draw[draw=#3,fill=#3!14,rounded corners=1.5pt,line width=0.5pt] (#1-0.33,#2-0.31) rectangle (#1+0.33,#2+0.31);
  \fill[#3] (#1-0.25,#2-0.24) rectangle ++(0.1,#4);
  \fill[#3] (#1-0.115,#2-0.24) rectangle ++(0.1,#5);
  \fill[#3] (#1+0.02,#2-0.24) rectangle ++(0.1,#6);
  \fill[#3] (#1+0.155,#2-0.24) rectangle ++(0.1,#7);
}
% the answers of one belief form one group
\newcommand{\ansgroup}[4]{% y, colour, bars of answer 1, bars of answer 2
  \draw[draw=bgedge!40,fill=white,rounded corners=3pt,line width=0.5pt] (\ax+0.2,#1-0.5) rectangle (\ax+3.2,#1+0.5);
  \node[text=#2,font=\normalsize] at (\ax+0.6,#1) {\faUserCircle};
  \ans{\ax+1.35}{#1}{#2}#3
  \ans{\ax+2.13}{#1}{#2}#4
  \node at (\ax+2.83,#1) {$\cdots$};
}
% advantages: scores relative to the group mean
\newcommand{\advchart}[5]{% name, y, title, group, bars
  \draw[bgedge!60,line width=0.4pt] (\gx+0.55,#2) -- (\gx+2.75,#2);
  \foreach \i/\h in {#5}
    {\pgfmathparse{\h>0}\ifnum\pgfmathresult=1
       \fill[bgup] (\gx+0.7+\i*0.33,#2) rectangle ++(0.22,\h);
     \else
       \fill[bgdown] (\gx+0.7+\i*0.33,#2) rectangle ++(0.22,\h);
     \fi}
  \node[align=center,inner sep=1pt,anchor=north] at (\gx+1.65,#2-0.48)
    {\textbf{#3}\\[-2pt]{\scriptsize #4}};
  \coordinate (#1) at (\gx,#2-0.2);
}
\begin{figure*}[t]
\centering
\resizebox{\textwidth}{!}{%
\begin{tikzpicture}
\def\bx{5.1}\def\ax{12.4}\def\sx{16.7}\def\gx{21.9}
\def\ytop{1.3}\def\ybot{-5.25}

% ---------------- background panels and step titles ----------------
\begin{scope}[on background layer]
  \draw[dashed,draw=black!55,line width=0.6pt,rounded corners=6pt] (\bx-0.3,\ytop) rectangle (\bx+5.85,\ybot);
  \draw[dashed,draw=black!55,line width=0.6pt,rounded corners=6pt] (\ax,\ytop) rectangle (\ax+3.4,\ybot);
  \draw[dashed,draw=black!55,line width=0.6pt,rounded corners=6pt] (\sx,\ytop) rectangle (\sx+4.3,\ybot);
  \draw[dashed,draw=black!55,line width=0.6pt,rounded corners=6pt] (\gx,\ytop) rectangle (\gx+3.3,\ybot);
\end{scope}
\node[num] (n1) at (\bx-0.05,\ytop+0.4) {1}; \node[ttl] at ($(n1.east)+(0.12,0)$) {Write belief lists};
\node[num] (n2) at (\ax+0.25,\ytop+0.4) {2}; \node[ttl] at ($(n2.east)+(0.12,0)$) {Answer as each belief};
\node[num] (n3) at (\sx+0.25,\ytop+0.4) {3}; \node[ttl] at ($(n3.east)+(0.12,0)$) {Credit and score};
\node[num] (n4) at (\gx+0.25,\ytop+0.4) {4}; \node[ttl] at ($(n4.east)+(0.12,0)$) {Advantages};

% ---------------- inputs ----------------
\draw[draw=bgedge,fill=white,rounded corners=3pt,line width=0.6pt] (-2.05,\ytop) rectangle (1.65,-2.65);
\node[inner sep=0pt,anchor=north] (img) at (-0.2,\ytop-0.2) {\includegraphics[width=3.3cm]{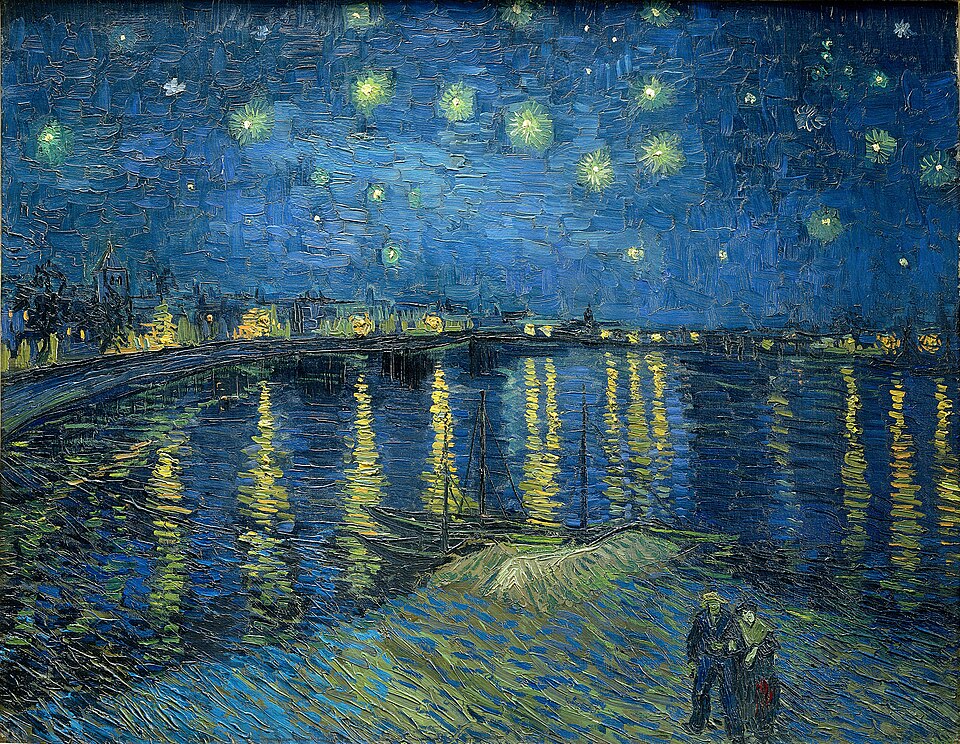}};
\node[align=center,inner sep=0pt,anchor=north] at ($(img.south)+(0,-0.12)$)
  {\textbf{Question}\\[0pt]{\footnotesize What does this painting}\\[-2pt]{\footnotesize make you feel?}};
\draw[draw=yB,fill=fY,rounded corners=3pt,line width=0.6pt] (-2.05,-3.15) rectangle (1.65,\ybot);
\foreach \i/\f in {0/\fa,1/\fb,2/\fa,3/\fd}
  {\node[text=bgedge!85,font=\large] at (-1.25+\i*0.7,-3.58) {\faUser};
   \node[font=\large,inner sep=0pt] at (-1.25+\i*0.7,-4.12) {\f};}
\node[inner sep=0pt] at (-0.2,-4.75) {{\color{bD}\faEye}\ \,\textbf{Observed annotations}};

% ---------------- the one trained policy ----------------
\node[policy,minimum width=16mm,minimum height=65.5mm] (pol) at (3.1,{(\ytop+\ybot)/2})
  {{\fontsize{24}{24}\selectfont\color{bgedge}\faRobot}\\[5pt]Policy\\Model};
\node[text=bgfire,font=\LARGE] at ($(pol.north east)+(-0.02,-0.02)$) {\faFire};
\draw[arr] (1.65,-0.225) -- (pol.west |- 0,-0.225);
\draw[arr] (1.65,-4.2) -- (pol.west |- 0,-4.2);

% ---------------- 1: belief lists ----------------
\cardstack{L}{0.95}{-1.4}
\node[anchor=west,inner sep=0pt] at (\bx+0.18,0.62) {\textbf{\faListUl\ \,List}};
\node[anchor=east,inner sep=0pt] at (\bx+5.15,0.62) {$\times L$};
\brow{0.05}{bA}{0.44}{55\%}{a calm night by the river}
\brow{-0.47}{bB}{0.25}{31\%}{the same stars over a far home}
\brow{-0.99}{bC}{0.11}{14\%}{the night sky is sublime}

\cardstack{P}{-2.1}{-5.0}
\node[anchor=west,inner sep=0pt] at (\bx+0.18,-2.43) {\textbf{{\color{bD}\faEye}\ \,Posterior list}};
\node[anchor=east,inner sep=0pt] at (\bx+5.15,-2.43) {$\times H$};
\fill[fD,rounded corners=2pt] (\bx+0.08,-4.84) rectangle (\bx+5.22,-4.32);
\brow{-3.02}{bA}{0.41}{51\%}{a calm night by the river}
\brow{-3.54}{bB}{0.23}{29\%}{the same stars over a far home}
\brow{-4.06}{bC}{0.10}{13\%}{the night sky is sublime}
\brow{-4.58}{bD}{0.06}{\textbf{7\%}}{\textbf{the dark water is unsettling}}

\draw[flow] (pol.east |- 0,-0.225) -- node[above,font=\scriptsize,inner sep=2pt] {Sample} (\bx-0.35,-0.225);
\draw[flow] (pol.east |- 0,-3.55) -- node[above,font=\scriptsize,inner sep=2pt] {Sample} (\bx-0.35,-3.55);

% ---------------- 2: answers of each belief ----------------
\ansgroup{0.55}{bA}{{0.46}{0.10}{0.12}{0.04}}{{0.40}{0.14}{0.10}{0.05}}
\ansgroup{-0.6}{bB}{{0.10}{0.46}{0.08}{0.05}}{{0.14}{0.40}{0.10}{0.06}}
\ansgroup{-1.75}{bC}{{0.12}{0.08}{0.44}{0.06}}{{0.10}{0.12}{0.42}{0.05}}
\node at (\ax+1.7,-2.45) {$\vdots$};
\ansgroup{-3.3}{bD}{{0.05}{0.12}{0.10}{0.44}}{{0.06}{0.10}{0.14}{0.40}}
% key: one answer, enlarged, with the four options under its bars
\draw[draw=bgedge!55,fill=black!4,rounded corners=3pt,line width=0.5pt] (\ax+0.2,-3.98) rectangle (\ax+3.2,-5.08);
\foreach \i/\h/\f in {0/0.42/\fa,1/0.13/\fb,2/0.17/\fc,3/0.07/\fd}
  {\fill[bgedge!60] (\ax+1.6+\i*0.4,-4.58) rectangle ++(0.2,\h);
   \node[font=\small,inner sep=0pt] at (\ax+1.7+\i*0.4,-4.81) {\f};}
\node[font=\scriptsize,text=bgedge,align=center,inner sep=0pt] at (\ax+0.85,-4.53) {one of the\\[-1pt]$n$ answers};
\draw[flow] (\bx+5.9,-1.975) -- (\ax-0.05,-1.975);
\node[align=center,font=\scriptsize,inner sep=1pt] at ({(\bx+5.85+\ax)/2},-1.15)
  {{\normalsize\color{bgedge}\faRobot}{\footnotesize\color{bgfire}\faFire}\\[1pt]same\\policy};

% ---------------- 3: credit, then scores ----------------
% held-out viewers and what each of them chose
\draw[draw=yB,fill=fY,rounded corners=3pt,line width=0.6pt] (\sx+0.3,1.05) rectangle (\sx+4.0,-0.34);
\node[inner sep=0pt] at (\sx+2.15,0.76) {\textbf{Held-out annotations}};
\foreach \i/\f in {0/\fa,1/\fb,2/\fa,3/\fd,4/\fa,5/\fc}
  {\node[text=black!55,font=\normalsize] at (\sx+0.9+\i*0.5,0.36) {\faUser};
   \node[font=\normalsize,inner sep=0pt] at (\sx+0.9+\i*0.5,-0.06) {\f};}
% Bayes' rule: each viewer is credited to the belief that likely produced the choice
\foreach \i/\c/\b in {0/bA/0,1/bB/1,2/bA/0,3/bD/3,4/bA/0,5/bC/2}
  \draw[\c,line width=1pt,line cap=round] (\sx+0.9+\i*0.5,-0.4) -- (\sx+0.95+\b*0.8,-1.28);
\foreach \b/\c in {0/bA,1/bB,2/bC,3/bD}
  \node[text=\c,font=\large,inner sep=0pt] at (\sx+0.95+\b*0.8,-1.5) {\faUserCircle};
\node[font=\scriptsize,inner sep=0pt] at (\sx+2.15,-1.93) {credit by Bayes' rule};

\tikzset{chip/.append style={minimum height=8.5mm}}
\node[chip,fill=fC,draw=bC] (sA) at (\sx+2.15,-2.72) {};
\node[chip,fill=fA,draw=bA] (sL) at (\sx+2.15,-3.67) {};
\node[chip,fill=fD,draw=bD] (sP) at (\sx+2.15,-4.62) {};
\foreach \n/\ic/\t/\s in {sA/\faBullseye/Answer score/by the belief's own viewers,
                          sL/\faBalanceScale/List score/by all held-out viewers,
                          sP/{\color{bD}\faEye}/Posterior score/fit to observed annotations}
  {\node[font=\normalsize,text=bgedge] at ($(\n.west)+(0.33,0)$) {\ic};
   \node[align=left,anchor=west,inner sep=0pt] at ($(\n.west)+(0.64,0)$) {\textbf{\t}\\[-2pt]{\scriptsize \s}};}
\draw[arr] (\sx+2.15,-2.08) -- (sA.north);
\draw[arr,rounded corners=3pt] (\sx+0.3,0.3) -- ++(-0.15,0) |- (sL.west);
\draw[flow] (\ax+3.45,-1.975) -- (\sx-0.05,-1.975);

% ---------------- 4: advantages within groups ----------------
\advchart{gA}{0.3}{Answer advantages}{answers of one belief}{0/0.36,1/-0.3,2/0.14,3/-0.1,4/-0.36,5/0.28}
\advchart{gL}{-1.85}{List advantages}{all $L{+}H$ lists}{0/-0.2,1/-0.34,2/0.1,3/-0.12,4/0.44,5/0.24}
\advchart{gP}{-4.0}{Posterior advantages}{the $H$ posterior lists}{0/-0.14,1/0.4,2/-0.38,3/0.2,4/0.08,5/-0.24}
\draw[arr] (sA.east) to[out=0,in=180] (gA);
\draw[arr] (sL.east) to[out=0,in=180] (gL);
\draw[arr] (sP.east) to[out=0,in=180] (gP);

% ---------------- one optimization step updates the same policy ----------------
\draw[flow,rounded corners=8pt] (\gx+3.35,-1.975) -- ++(0.5,0) |- (10,\ytop+1.6) -| (pol.north);
\node[above,font=\normalsize] at (13.4,\ytop+1.6) {\textbf{Policy Optimization}\ \ {\small all advantages update the one policy}};
\end{tikzpicture}}%
\caption{\textbf{BGPO training on a single question.}
(1) A shared policy generates $L$ belief lists from the question and $H$ posterior lists conditioned on observed annotations.
(2) It samples $n$ answers per belief.
(3) Bayes' rule assigns soft credit for held-out annotations to beliefs, yielding belief-specific answer scores. List scores use all held-out annotations, and posterior scores use observed annotations.
(4) Within-group advantages update the shared policy.}
\label{fig:method}
\end{figure*}

%% file: fig/cases.tex
% Case studies: generated by BGPO/examples/case_study/make_cases_fig.py. Do not edit numbers by hand.
% Colours are sampled from the reference figure supplied by the authors.
\definecolor{cscream}{HTML}{FDF7E5}
\definecolor{csgreen}{HTML}{E6FCEB}
\definecolor{csblue}{HTML}{E0F7FD}
\definecolor{csb1}{HTML}{9CD8FC}
\definecolor{csb2}{HTML}{FCC048}
\definecolor{csb3}{HTML}{FCCCCC}
\definecolor{csb4}{HTML}{F6E4F0}
\definecolor{csnavy}{HTML}{004878}
\definecolor{csedge}{HTML}{2A2A2A}
\begin{figure*}[t]
\centering
\resizebox{\textwidth}{!}{%
\begin{tikzpicture}[
  cspanel/.style={draw=csedge,line width=0.6pt,rounded corners=3pt},
  cstitle/.style={draw=csedge,fill=white,line width=0.5pt,rounded corners=2pt,font=\small,inner sep=3pt},
  cshead/.style={font=\scriptsize\ttfamily\bfseries,text=csedge,inner sep=0pt},
  csbox/.style={draw=csedge,fill=white,line width=0.5pt,rounded corners=3pt},
  cschip/.style={font=\scriptsize,execute at begin node={\hyphenpenalty=10000 \exhyphenpenalty=10000 },draw=csedge,line width=0.5pt,rounded corners=3pt,align=left,minimum height=0.9cm,inner sep=2.5pt},
  cstext/.style={align=left,inner sep=0pt,execute at begin node={\hyphenpenalty=10000 \exhyphenpenalty=10000 }},
  cstiny/.style={font=\tiny,inner sep=0pt,text=csedge},
  cstrack/.style={draw=none,fill=white},
]
\draw[cspanel,fill=cscream] (0.0,0) rectangle (5.75,-9.0);
\node[cstitle] at (2.875,0.36) {\textit{A) Landscape, Thomas Cole}};
\node[cshead] at (2.875,-0.33999999999999997) {Question};
\draw[csbox] (0.15,-0.56) rectangle (5.6,-2.71);
\node[inner sep=0pt] at (1.45,-1.635) {\includegraphics[width=2.35cm,height=1.95cm,keepaspectratio]{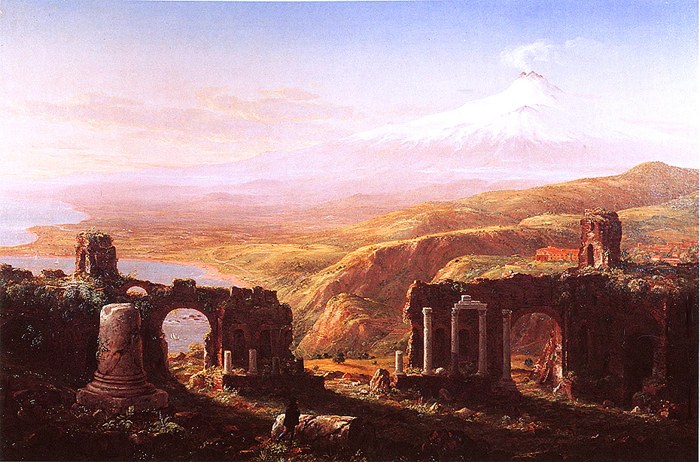}};
\node[cstext,font=\scriptsize,text width=2.65cm,anchor=west] at (2.75,-1.635) {\textbf{Q:} Which emotion does this painting evoke?\\[2pt]\textbf{Options:} nine emotions\\[2pt]\textbf{Label:} annotation language};
\node[cshead] at (2.875,-2.99) {Beliefs written by BGPO};
\node[cschip,fill=csb1,minimum width=3.55cm,text width=3.3499999999999996cm] at (1.9249999999999998,-3.81) {It makes people wonder about the history of the ancient city.};
\draw[cstrack] (3.8499999999999996,-3.61) rectangle ++(1.05,0.17);
\fill[csb1] (3.8499999999999996,-3.61) rectangle ++(0.379,0.17);
\draw[csedge,line width=0.4pt] (3.8499999999999996,-3.61) rectangle ++(1.05,0.17);
\node[cstiny,anchor=west] at (4.919999999999999,-3.525) {\textbf{36\%}};
\node[cstiny,anchor=west] at (3.82,-3.83) {picks \textbf{awe}};
\node[cstiny,anchor=west] at (3.82,-4.08) {45\% Arabic};
\node[cschip,fill=csb2,minimum width=3.55cm,text width=3.3499999999999996cm] at (1.9249999999999998,-5.01) {It brings a sense of peace while looking over a beautiful landscape.};
\draw[cstrack] (3.8499999999999996,-4.81) rectangle ++(1.05,0.17);
\fill[csb2] (3.8499999999999996,-4.81) rectangle ++(0.283,0.17);
\draw[csedge,line width=0.4pt] (3.8499999999999996,-4.81) rectangle ++(1.05,0.17);
\node[cstiny,anchor=west] at (4.919999999999999,-4.725) {\textbf{27\%}};
\node[cstiny,anchor=west] at (3.82,-5.029999999999999) {picks \textbf{contentment}};
\node[cstiny,anchor=west] at (3.82,-5.279999999999999) {47\% Chinese};
\node[cschip,fill=csb3,minimum width=3.55cm,text width=3.3499999999999996cm] at (1.9249999999999998,-6.21) {It makes viewers feel the power of nature erasing human achievements.};
\draw[cstrack] (3.8499999999999996,-6.01) rectangle ++(1.05,0.17);
\fill[csb3] (3.8499999999999996,-6.01) rectangle ++(0.389,0.17);
\draw[csedge,line width=0.4pt] (3.8499999999999996,-6.01) rectangle ++(1.05,0.17);
\node[cstiny,anchor=west] at (4.919999999999999,-5.925) {\textbf{37\%}};
\node[cstiny,anchor=west] at (3.82,-6.2299999999999995) {picks \textbf{sadness}};
\node[cstiny,anchor=west] at (3.82,-6.48) {37\% Chinese};
\node[cshead] at (2.875,-7.07) {Comparison of Prediction and Label};
\node[cstiny] at (2.058333333333333,-7.410000000000001) {awe};
\node[cstiny] at (3.4749999999999996,-7.410000000000001) {contentment};
\node[cstiny] at (4.891666666666667,-7.410000000000001) {sadness};
\node[cstiny,anchor=east] at (1.2699999999999998,-7.755000000000001) {English};
\draw[cstrack] (1.42,-7.845000000000001) rectangle ++(1.2766666666666668,0.18);
\fill[csb1] (1.42,-7.845000000000001) rectangle ++(0.386,0.18);
\draw[csedge,line width=0.4pt] (1.42,-7.845000000000001) rectangle ++(1.2766666666666668,0.18);
\draw[csnavy,line width=1.3pt] (2.058,-7.8950000000000005) -- ++(0,0.28);
\draw[cstrack] (2.8366666666666664,-7.845000000000001) rectangle ++(1.2766666666666668,0.18);
\fill[csb2] (2.8366666666666664,-7.845000000000001) rectangle ++(0.271,0.18);
\draw[csedge,line width=0.4pt] (2.8366666666666664,-7.845000000000001) rectangle ++(1.2766666666666668,0.18);
\draw[csnavy,line width=1.3pt] (2.837,-7.8950000000000005) -- ++(0,0.28);
\draw[cstrack] (4.253333333333334,-7.845000000000001) rectangle ++(1.2766666666666668,0.18);
\fill[csb3] (4.253333333333334,-7.845000000000001) rectangle ++(0.279,0.18);
\draw[csedge,line width=0.4pt] (4.253333333333334,-7.845000000000001) rectangle ++(1.2766666666666668,0.18);
\draw[csnavy,line width=1.3pt] (4.253,-7.8950000000000005) -- ++(0,0.28);
\node[cstiny,anchor=east] at (1.2699999999999998,-8.085) {Arabic};
\draw[cstrack] (1.42,-8.175) rectangle ++(1.2766666666666668,0.18);
\fill[csb1] (1.42,-8.175) rectangle ++(0.521,0.18);
\draw[csedge,line width=0.4pt] (1.42,-8.175) rectangle ++(1.2766666666666668,0.18);
\draw[csnavy,line width=1.3pt] (2.441,-8.225000000000001) -- ++(0,0.28);
\draw[cstrack] (2.8366666666666664,-8.175) rectangle ++(1.2766666666666668,0.18);
\fill[csb2] (2.8366666666666664,-8.175) rectangle ++(0.264,0.18);
\draw[csedge,line width=0.4pt] (2.8366666666666664,-8.175) rectangle ++(1.2766666666666668,0.18);
\draw[csnavy,line width=1.3pt] (2.837,-8.225000000000001) -- ++(0,0.28);
\draw[cstrack] (4.253333333333334,-8.175) rectangle ++(1.2766666666666668,0.18);
\fill[csb3] (4.253333333333334,-8.175) rectangle ++(0.199,0.18);
\draw[csedge,line width=0.4pt] (4.253333333333334,-8.175) rectangle ++(1.2766666666666668,0.18);
\draw[csnavy,line width=1.3pt] (4.253,-8.225000000000001) -- ++(0,0.28);
\node[cstiny,anchor=east] at (1.2699999999999998,-8.415) {Chinese};
\draw[cstrack] (1.42,-8.504999999999999) rectangle ++(1.2766666666666668,0.18);
\fill[csb1] (1.42,-8.504999999999999) rectangle ++(0.256,0.18);
\draw[csedge,line width=0.4pt] (1.42,-8.504999999999999) rectangle ++(1.2766666666666668,0.18);
\draw[csnavy,line width=1.3pt] (1.420,-8.555) -- ++(0,0.28);
\draw[cstrack] (2.8366666666666664,-8.504999999999999) rectangle ++(1.2766666666666668,0.18);
\fill[csb2] (2.8366666666666664,-8.504999999999999) rectangle ++(0.439,0.18);
\draw[csedge,line width=0.4pt] (2.8366666666666664,-8.504999999999999) rectangle ++(1.2766666666666668,0.18);
\draw[csnavy,line width=1.3pt] (4.113,-8.555) -- ++(0,0.28);
\draw[cstrack] (4.253333333333334,-8.504999999999999) rectangle ++(1.2766666666666668,0.18);
\fill[csb3] (4.253333333333334,-8.504999999999999) rectangle ++(0.267,0.18);
\draw[csedge,line width=0.4pt] (4.253333333333334,-8.504999999999999) rectangle ++(1.2766666666666668,0.18);
\draw[csnavy,line width=1.3pt] (4.253,-8.555) -- ++(0,0.28);
\fill[csb1] (1.425,-8.870000000000001) rectangle ++(0.4,0.14);\draw[csedge,line width=0.4pt] (1.425,-8.870000000000001) rectangle ++(0.4,0.14);
\node[cstiny,anchor=west] at (1.905,-8.8) {BGPO};
\draw[csnavy,line width=1.3pt] (2.875,-8.92) -- ++(0,0.24);
\node[cstiny,anchor=west] at (2.9450000000000003,-8.8) {human};
\draw[cspanel,fill=csgreen] (5.97,0) rectangle (11.719999999999999,-9.0);
\node[cstitle] at (8.844999999999999,0.36) {\textit{B) Animal portrait, George Stubbs}};
\node[cshead] at (8.844999999999999,-0.33999999999999997) {Question};
\draw[csbox] (6.12,-0.56) rectangle (11.569999999999999,-2.71);
\node[inner sep=0pt] at (7.42,-1.635) {\includegraphics[width=2.35cm,height=1.95cm,keepaspectratio]{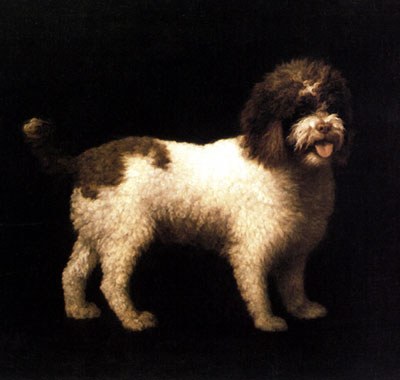}};
\node[cstext,font=\scriptsize,text width=2.65cm,anchor=west] at (8.72,-1.635) {\textbf{Q:} Which emotion does this painting evoke?\\[2pt]\textbf{Options:} nine emotions\\[2pt]\textbf{Label:} annotation language};
\node[cshead] at (8.844999999999999,-2.99) {Beliefs written by BGPO};
\node[cschip,fill=csb1,minimum width=3.55cm,text width=3.3499999999999996cm] at (7.895,-3.81) {It makes people want to get closer to their pets.};
\draw[cstrack] (9.82,-3.61) rectangle ++(1.05,0.17);
\fill[csb1] (9.82,-3.61) rectangle ++(0.427,0.17);
\draw[csedge,line width=0.4pt] (9.82,-3.61) rectangle ++(1.05,0.17);
\node[cstiny,anchor=west] at (10.89,-3.525) {\textbf{41\%}};
\node[cstiny,anchor=west] at (9.790000000000001,-3.83) {picks \textbf{contentment}};
\node[cstiny,anchor=west] at (9.790000000000001,-4.08) {49\% Chinese};
\node[cschip,fill=csb2,minimum width=3.55cm,text width=3.3499999999999996cm] at (7.895,-5.01) {It brings back great memories of childhood adventures with dogs.};
\draw[cstrack] (9.82,-4.81) rectangle ++(1.05,0.17);
\fill[csb2] (9.82,-4.81) rectangle ++(0.336,0.17);
\draw[csedge,line width=0.4pt] (9.82,-4.81) rectangle ++(1.05,0.17);
\node[cstiny,anchor=west] at (10.89,-4.725) {\textbf{32\%}};
\node[cstiny,anchor=west] at (9.790000000000001,-5.029999999999999) {picks \textbf{amusement}};
\node[cstiny,anchor=west] at (9.790000000000001,-5.279999999999999) {43\% Arabic};
\node[cschip,fill=csb3,minimum width=3.55cm,text width=3.3499999999999996cm] at (7.895,-6.21) {It makes people wonder who actually painted this masterpiece.};
\draw[cstrack] (9.82,-6.01) rectangle ++(1.05,0.17);
\fill[csb3] (9.82,-6.01) rectangle ++(0.287,0.17);
\draw[csedge,line width=0.4pt] (9.82,-6.01) rectangle ++(1.05,0.17);
\node[cstiny,anchor=west] at (10.89,-5.925) {\textbf{27\%}};
\node[cstiny,anchor=west] at (9.790000000000001,-6.2299999999999995) {picks \textbf{excitement}};
\node[cstiny,anchor=west] at (9.790000000000001,-6.48) {39\% English};
\node[cshead] at (8.844999999999999,-7.07) {Comparison of Prediction and Label};
\node[cstiny] at (8.028333333333334,-7.410000000000001) {contentment};
\node[cstiny] at (9.445,-7.410000000000001) {amusement};
\node[cstiny] at (10.861666666666668,-7.410000000000001) {excitement};
\node[cstiny,anchor=east] at (7.24,-7.755000000000001) {English};
\draw[cstrack] (7.390000000000001,-7.845000000000001) rectangle ++(1.2766666666666668,0.18);
\fill[csb1] (7.390000000000001,-7.845000000000001) rectangle ++(0.381,0.18);
\draw[csedge,line width=0.4pt] (7.390000000000001,-7.845000000000001) rectangle ++(1.2766666666666668,0.18);
\draw[csnavy,line width=1.3pt] (7.901,-7.8950000000000005) -- ++(0,0.28);
\draw[cstrack] (8.806666666666667,-7.845000000000001) rectangle ++(1.2766666666666668,0.18);
\fill[csb2] (8.806666666666667,-7.845000000000001) rectangle ++(0.288,0.18);
\draw[csedge,line width=0.4pt] (8.806666666666667,-7.845000000000001) rectangle ++(1.2766666666666668,0.18);
\draw[csnavy,line width=1.3pt] (9.062,-7.8950000000000005) -- ++(0,0.28);
\draw[cstrack] (10.223333333333334,-7.845000000000001) rectangle ++(1.2766666666666668,0.18);
\fill[csb3] (10.223333333333334,-7.845000000000001) rectangle ++(0.130,0.18);
\draw[csedge,line width=0.4pt] (10.223333333333334,-7.845000000000001) rectangle ++(1.2766666666666668,0.18);
\draw[csnavy,line width=1.3pt] (10.479,-7.8950000000000005) -- ++(0,0.28);
\node[cstiny,anchor=east] at (7.24,-8.085) {Arabic};
\draw[cstrack] (7.390000000000001,-8.175) rectangle ++(1.2766666666666668,0.18);
\fill[csb1] (7.390000000000001,-8.175) rectangle ++(0.374,0.18);
\draw[csedge,line width=0.4pt] (7.390000000000001,-8.175) rectangle ++(1.2766666666666668,0.18);
\draw[csnavy,line width=1.3pt] (7.390,-8.225000000000001) -- ++(0,0.28);
\draw[cstrack] (8.806666666666667,-8.175) rectangle ++(1.2766666666666668,0.18);
\fill[csb2] (8.806666666666667,-8.175) rectangle ++(0.324,0.18);
\draw[csedge,line width=0.4pt] (8.806666666666667,-8.175) rectangle ++(1.2766666666666668,0.18);
\draw[csnavy,line width=1.3pt] (9.573,-8.225000000000001) -- ++(0,0.28);
\draw[cstrack] (10.223333333333334,-8.175) rectangle ++(1.2766666666666668,0.18);
\fill[csb3] (10.223333333333334,-8.175) rectangle ++(0.113,0.18);
\draw[csedge,line width=0.4pt] (10.223333333333334,-8.175) rectangle ++(1.2766666666666668,0.18);
\draw[csnavy,line width=1.3pt] (10.223,-8.225000000000001) -- ++(0,0.28);
\node[cstiny,anchor=east] at (7.24,-8.415) {Chinese};
\draw[cstrack] (7.390000000000001,-8.504999999999999) rectangle ++(1.2766666666666668,0.18);
\fill[csb1] (7.390000000000001,-8.504999999999999) rectangle ++(0.709,0.18);
\draw[csedge,line width=0.4pt] (7.390000000000001,-8.504999999999999) rectangle ++(1.2766666666666668,0.18);
\draw[csnavy,line width=1.3pt] (8.028,-8.555) -- ++(0,0.28);
\draw[cstrack] (8.806666666666667,-8.504999999999999) rectangle ++(1.2766666666666668,0.18);
\fill[csb2] (8.806666666666667,-8.504999999999999) rectangle ++(0.120,0.18);
\draw[csedge,line width=0.4pt] (8.806666666666667,-8.504999999999999) rectangle ++(1.2766666666666668,0.18);
\draw[csnavy,line width=1.3pt] (8.807,-8.555) -- ++(0,0.28);
\draw[cstrack] (10.223333333333334,-8.504999999999999) rectangle ++(1.2766666666666668,0.18);
\fill[csb3] (10.223333333333334,-8.504999999999999) rectangle ++(0.111,0.18);
\draw[csedge,line width=0.4pt] (10.223333333333334,-8.504999999999999) rectangle ++(1.2766666666666668,0.18);
\draw[csnavy,line width=1.3pt] (10.862,-8.555) -- ++(0,0.28);
\fill[csb1] (7.394999999999999,-8.870000000000001) rectangle ++(0.4,0.14);\draw[csedge,line width=0.4pt] (7.394999999999999,-8.870000000000001) rectangle ++(0.4,0.14);
\node[cstiny,anchor=west] at (7.874999999999998,-8.8) {BGPO};
\draw[csnavy,line width=1.3pt] (8.844999999999999,-8.92) -- ++(0,0.24);
\node[cstiny,anchor=west] at (8.915,-8.8) {human};
\draw[cspanel,fill=csblue] (11.94,0) rectangle (17.689999999999998,-9.0);
\node[cstitle] at (14.815,0.36) {\textit{C) Engraving, Gustave Dor\'e}};
\node[cshead] at (14.815,-0.33999999999999997) {Question};
\draw[csbox] (12.09,-0.56) rectangle (17.54,-2.71);
\node[inner sep=0pt] at (13.39,-1.635) {\includegraphics[width=2.35cm,height=1.95cm,keepaspectratio]{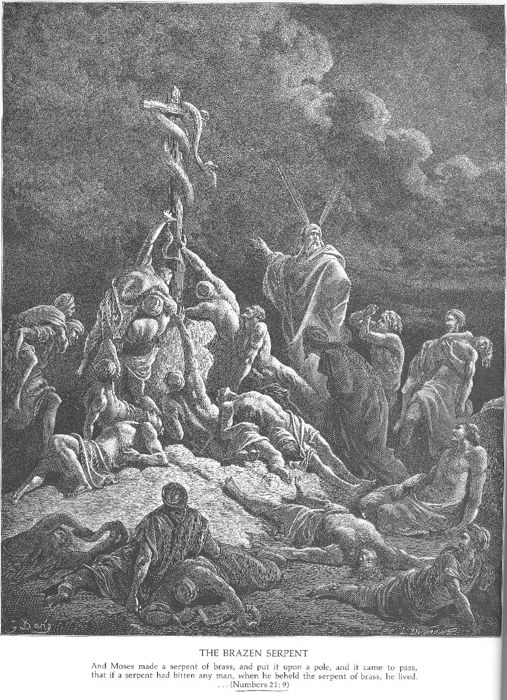}};
\node[cstext,font=\scriptsize,text width=2.65cm,anchor=west] at (14.69,-1.635) {\textbf{Q:} Which emotion does this painting evoke?\\[2pt]\textbf{Options:} nine emotions\\[2pt]\textbf{Label:} annotation language};
\node[cshead] at (14.815,-2.99) {Beliefs written by BGPO};
\node[cschip,fill=csb1,minimum width=3.55cm,text width=3.3499999999999996cm] at (13.865,-3.81) {It makes people terrified of the unknown dangers in the ancient world.};
\draw[cstrack] (15.790000000000001,-3.61) rectangle ++(1.05,0.17);
\fill[csb1] (15.790000000000001,-3.61) rectangle ++(0.489,0.17);
\draw[csedge,line width=0.4pt] (15.790000000000001,-3.61) rectangle ++(1.05,0.17);
\node[cstiny,anchor=west] at (16.86,-3.525) {\textbf{47\%}};
\node[cstiny,anchor=west] at (15.760000000000002,-3.83) {picks \textbf{fear}};
\node[cstiny,anchor=west] at (15.760000000000002,-4.08) {42\% Chinese};
\node[cschip,fill=csb2,minimum width=3.55cm,text width=3.3499999999999996cm] at (13.865,-5.01) {It brings a deep sense of respect for the historical importance of religious stories.};
\draw[cstrack] (15.790000000000001,-4.81) rectangle ++(1.05,0.17);
\fill[csb2] (15.790000000000001,-4.81) rectangle ++(0.180,0.17);
\draw[csedge,line width=0.4pt] (15.790000000000001,-4.81) rectangle ++(1.05,0.17);
\node[cstiny,anchor=west] at (16.86,-4.725) {\textbf{17\%}};
\node[cstiny,anchor=west] at (15.760000000000002,-5.029999999999999) {picks \textbf{awe}};
\node[cstiny,anchor=west] at (15.760000000000002,-5.279999999999999) {41\% Arabic};
\node[cschip,fill=csb3,minimum width=3.55cm,text width=3.3499999999999996cm] at (13.865,-6.21) {It evokes a strong sense of sadness because it shows how people punish their own neighbors.};
\draw[cstrack] (15.790000000000001,-6.01) rectangle ++(1.05,0.17);
\fill[csb3] (15.790000000000001,-6.01) rectangle ++(0.380,0.17);
\draw[csedge,line width=0.4pt] (15.790000000000001,-6.01) rectangle ++(1.05,0.17);
\node[cstiny,anchor=west] at (16.86,-5.925) {\textbf{36\%}};
\node[cstiny,anchor=west] at (15.760000000000002,-6.2299999999999995) {picks \textbf{sadness}};
\node[cstiny,anchor=west] at (15.760000000000002,-6.48) {36\% English};
\node[cshead] at (14.815,-7.07) {Comparison of Prediction and Label};
\node[cstiny] at (13.998333333333333,-7.410000000000001) {fear};
\node[cstiny] at (15.415,-7.410000000000001) {awe};
\node[cstiny] at (16.831666666666667,-7.410000000000001) {sadness};
\node[cstiny,anchor=east] at (13.209999999999999,-7.755000000000001) {English};
\draw[cstrack] (13.36,-7.845000000000001) rectangle ++(1.2766666666666668,0.18);
\fill[csb1] (13.36,-7.845000000000001) rectangle ++(0.470,0.18);
\draw[csedge,line width=0.4pt] (13.36,-7.845000000000001) rectangle ++(1.2766666666666668,0.18);
\draw[csnavy,line width=1.3pt] (13.871,-7.8950000000000005) -- ++(0,0.28);
\draw[cstrack] (14.776666666666666,-7.845000000000001) rectangle ++(1.2766666666666668,0.18);
\fill[csb2] (14.776666666666666,-7.845000000000001) rectangle ++(0.142,0.18);
\draw[csedge,line width=0.4pt] (14.776666666666666,-7.845000000000001) rectangle ++(1.2766666666666668,0.18);
\draw[csnavy,line width=1.3pt] (15.032,-7.8950000000000005) -- ++(0,0.28);
\draw[cstrack] (16.19333333333333,-7.845000000000001) rectangle ++(1.2766666666666668,0.18);
\fill[csb3] (16.19333333333333,-7.845000000000001) rectangle ++(0.289,0.18);
\draw[csedge,line width=0.4pt] (16.19333333333333,-7.845000000000001) rectangle ++(1.2766666666666668,0.18);
\draw[csnavy,line width=1.3pt] (16.449,-7.8950000000000005) -- ++(0,0.28);
\node[cstiny,anchor=east] at (13.209999999999999,-8.085) {Arabic};
\draw[cstrack] (13.36,-8.175) rectangle ++(1.2766666666666668,0.18);
\fill[csb1] (13.36,-8.175) rectangle ++(0.416,0.18);
\draw[csedge,line width=0.4pt] (13.36,-8.175) rectangle ++(1.2766666666666668,0.18);
\draw[csnavy,line width=1.3pt] (13.615,-8.225000000000001) -- ++(0,0.28);
\draw[cstrack] (14.776666666666666,-8.175) rectangle ++(1.2766666666666668,0.18);
\fill[csb2] (14.776666666666666,-8.175) rectangle ++(0.201,0.18);
\draw[csedge,line width=0.4pt] (14.776666666666666,-8.175) rectangle ++(1.2766666666666668,0.18);
\draw[csnavy,line width=1.3pt] (15.287,-8.225000000000001) -- ++(0,0.28);
\draw[cstrack] (16.19333333333333,-8.175) rectangle ++(1.2766666666666668,0.18);
\fill[csb3] (16.19333333333333,-8.175) rectangle ++(0.275,0.18);
\draw[csedge,line width=0.4pt] (16.19333333333333,-8.175) rectangle ++(1.2766666666666668,0.18);
\draw[csnavy,line width=1.3pt] (16.193,-8.225000000000001) -- ++(0,0.28);
\node[cstiny,anchor=east] at (13.209999999999999,-8.415) {Chinese};
\draw[cstrack] (13.36,-8.504999999999999) rectangle ++(1.2766666666666668,0.18);
\fill[csb1] (13.36,-8.504999999999999) rectangle ++(0.557,0.18);
\draw[csedge,line width=0.4pt] (13.36,-8.504999999999999) rectangle ++(1.2766666666666668,0.18);
\draw[csnavy,line width=1.3pt] (14.381,-8.555) -- ++(0,0.28);
\draw[cstrack] (14.776666666666666,-8.504999999999999) rectangle ++(1.2766666666666668,0.18);
\fill[csb2] (14.776666666666666,-8.504999999999999) rectangle ++(0.120,0.18);
\draw[csedge,line width=0.4pt] (14.776666666666666,-8.504999999999999) rectangle ++(1.2766666666666668,0.18);
\draw[csnavy,line width=1.3pt] (14.777,-8.555) -- ++(0,0.28);
\draw[cstrack] (16.19333333333333,-8.504999999999999) rectangle ++(1.2766666666666668,0.18);
\fill[csb3] (16.19333333333333,-8.504999999999999) rectangle ++(0.248,0.18);
\draw[csedge,line width=0.4pt] (16.19333333333333,-8.504999999999999) rectangle ++(1.2766666666666668,0.18);
\draw[csnavy,line width=1.3pt] (16.193,-8.555) -- ++(0,0.28);
\fill[csb1] (13.365,-8.870000000000001) rectangle ++(0.4,0.14);\draw[csedge,line width=0.4pt] (13.365,-8.870000000000001) rectangle ++(0.4,0.14);
\node[cstiny,anchor=west] at (13.845,-8.8) {BGPO};
\draw[csnavy,line width=1.3pt] (14.815,-8.92) -- ++(0,0.24);
\node[cstiny,anchor=west] at (14.885,-8.8) {human};
\end{tikzpicture}}%
\caption{\textbf{Case studies.} Belief lists written by BGPO for three ArtELingo test paintings. Each belief shows its share, the option its holders pick most often, and the label most common among its holders. Bottom: the prediction for each label that results from mixing the beliefs (bar) against the choices of annotators with that label (tick).}
\label{fig:cases}
\end{figure*}

%% file: sec/supp_method.tex
% Supplementary material for BGPO: details moved out of the main text.
% Cross-references to the main paper are written as plain text ("Eq. (4) of the main paper").
% Main-paper equation numbers: (1) population model, (2) pooling-loss split, (3) rollout mixture \hat P_h,
% (4) responsibility r_k, (5) answer reward S, (6) list reward R, (7) posterior reward R_q, (8) clipped surrogate. The recovery rate is defined only here.
\clearpage
\section{Training objective}
\label{supp:objective}

\paragraph{Batch.}
A training batch is a set $\mathcal{O}$ of triples $(o,c,A)$: a sequence $o$ with $|o|$ tokens, the context $c$ it was sampled under, and one scalar advantage $A$. Three kinds of triples occur for each question $x$:
\begin{itemize}
  \item $(h,x,A(h))$ for every list $h$ in the mixed group $\mathcal{H}_x$ of the question, which holds its $L$ policy lists and its $H$ posterior lists;
  \item $(\hat h,(x,\Dobs),A_q(\hat h))$ for every posterior list, so a posterior list enters the batch twice, once to train the posterior and once to train the policy;
  \item $(z_{kj},(x,h,k),A(z_{kj}))$ for every answer.
\end{itemize}
Here $A(h)=R(h)-\frac{1}{L+H-1}\sum_{h'\in\mathcal{H}_x\setminus\{h\}}R(h')$ and $A_q(\hat h)=R_q(\hat h)-\frac{1}{H-1}\sum_{\hat h'\neq\hat h}R_q(\hat h')$, with $R$ and $R_q$ from Eqs.~(6) and (7) of the main paper, and $A(z_{kj})=S(z_{kj})-\frac{1}{n-1}\sum_{j'\neq j}S(z_{kj'})$ with $S$ from Eq.~(5) of the main paper.

\paragraph{Surrogate.}
Let $\thetaold$ be the parameters that sampled the batch, $s_t=\pi_\theta(o_t\mid c,o_{<t})/\pi_{\thetaold}(o_t\mid c,o_{<t})$ the token ratio, $\bar s_t=\clip(s_t,1-\epsilon_{\text{low}},1+\epsilon_{\text{high}})$ its clipped value, and $N_{\mathcal{O}}=\sum_{(o,c,A)\in\mathcal{O}}|o|$ the number of tokens in the batch. We maximize the clipped surrogate~\citep{schulman2017ppo} in the form used by GRPO and DAPO~\citep{shao2024deepseekmath,yu2025dapo},
\begin{equation}
  \mathcal{J}(\theta)=\frac{1}{N_{\mathcal{O}}}\sum_{(o,c,A)\in\mathcal{O}}\;\sum_{t=1}^{|o|}\min\bigl(s_tA,\;\bar s_tA\bigr).
  \label{supp:eq:surrogate}
\end{equation}
For a posterior list under the policy's prompt, $c=x$ and $\pi_{\thetaold}(\hat h_t\mid x,\hat h_{<t})$ is computed by teacher forcing, although $\hat h$ was sampled under $(x,\Dobs)$.

\paragraph{Why each choice.}
\begin{itemize}
  \item \emph{Token-level mean.} Averaging over all tokens in the batch, as DAPO does, gives a token the same weight whether it belongs to a list or an answer, and long sequences are not down-weighted~\citep{yu2025dapo}.
  \item \emph{Clip-higher.} With $\epsilon_{\text{high}}>\epsilon_{\text{low}}$, a low-probability token can gain more probability per update~\citep{yu2025dapo}. Under a symmetric $\epsilon=0.2$, a token at probability $0.01$ can reach at most $0.012$ in one update. Tokens that name a rare belief are of this kind.
  \item \emph{No standard-deviation normalization.} Following Dr.~GRPO~\citep{liu2025drgrpo}, advantages are not divided by their group's standard deviation. The answers for a rare belief all score near zero because $S$ is weighted by that belief's credit. Normalizing would give every belief's answers updates of the same size, whatever its credit, and so undo the responsibility weighting of Eq.~(4) of the main paper.
  \item \emph{No KL toward the base model.} As in DAPO~\citep{yu2025dapo}, the surrogate has no KL penalty toward the base model. The base model states a single pooled distribution, so such a penalty would pull lists back toward one pooled belief, the behavior BGPO exists to leave. The only KL term in BGPO is the one in $R_q$, which ties the posterior to the policy.
\end{itemize}

\paragraph{Posterior lists in the policy's update.}
At $\theta=\thetaold$, the gradient of the posterior lists' terms in \cref{supp:eq:surrogate} is $A(\hat h)\nabla_\theta\log\pi_\theta(\hat h\mid x)$ per list. In expectation over $\hat h\sim q_\theta(\cdot\mid x,\Dobs)$ this is the gradient of $\mathbb{E}_{\hat h\sim q_\theta}[A(\hat h)\log\pi_\theta(\hat h\mid x)]$: an advantage-weighted likelihood of posterior lists under the policy's prompt. A posterior list that predicts $\Deval$ better than its group on average becomes more likely under $x$, and one that predicts worse becomes less likely. The exact off-policy correction would multiply each term by $\pi_\theta(\hat h\mid x)/q_\theta(\hat h\mid x,\Dobs)$. This weight is smallest for lists that describe beliefs the policy lacks, so it would erase the guidance the posterior exists to give. LUFFY~\citep{yan2025luffy} also adds off-policy traces to GRPO groups. It sets their behavior probability to one, reshapes the resulting ratio with $f(x)=x/(x+\gamma)$, $\gamma=0.1$, and does not clip off-policy tokens. We instead keep the policy's own teacher-forced old probability, so posterior lists and the policy's own lists are clipped by the same trust region.

\section{Derivations}
\label{supp:derivations}

\paragraph{Pooling (Sec.~3 of the main paper).}
For a prediction $p$ that cannot depend on the person, $\mathbb{E}_{(y,g)}[\log p(y)]=\sum_yP(y\mid x)\log p(y)\le\sum_yP(y\mid x)\log P(y\mid x)$ by Gibbs' inequality, with equality iff $p=P(\cdot\mid x)$. A 0/1 reward for a sampled choice $\hat y\sim p$ has expectation $\sum_y p(y)P(y\mid x)$, which is linear in $p$ and maximized by a point mass on $\arg\max_yP(y\mid x)$.

\paragraph{The cost of ignoring evidence.}
By definition of mutual information, $\mathbb{E}\log P(y\mid b,x)-\mathbb{E}\log P(y\mid x)=I(Y;B\mid x)$ and $\mathbb{E}\log P(y\mid g,x)-\mathbb{E}\log P(y\mid x)=I(Y;G\mid x)$. The chain rule expands $I(Y;B,G\mid x)$ both as $I(Y;G\mid x)+I(Y;B\mid G,x)$ and as $I(Y;B\mid x)+I(Y;G\mid B,x)$. Since $Y\perp G\mid B,x$, $I(Y;G\mid B,x)=0$, hence
\begin{equation}
  I(Y;B\mid x)=I(Y;G\mid x)+I(Y;B\mid G,x).
\end{equation}
$I(Y;B\mid G,x)=0$ iff $P(y\mid b,g,x)=P(y\mid g,x)$ whenever $P(b,g\mid x)>0$; with $Y\perp G\mid B,x$ this says that all beliefs present within each label have the same $P(y\mid b,x)$.

\paragraph{Credit weights.}
For weights $\omega_k^m\ge0$ with $\Omega_k=\sum_m\omega_k^m>0$, $\sum_m\omega_k^m\log U_k(y_m)=\Omega_k\sum_y e_k(y)\log U_k(y)$ with $e_k(y)=\sum_m\omega_k^m\mathbf{1}[y_m=y]/\Omega_k$. By Gibbs' inequality the maximum over distributions $U_k$ is at $U_k=e_k$. For the Fisher identity, let $a_k=w_k\eta_k(g)\bar U_k(y)>0$; then $\nabla\log\sum_ka_k=\sum_k\nabla a_k/\sum_{k'}a_{k'}=\sum_k\frac{a_k}{\sum_{k'}a_{k'}}\nabla\log a_k=\sum_kr_k(y,g)\nabla\log a_k$.

\paragraph{Target of the answer reward.}
With $r_k$ held fixed, the expectation of Eq.~(5) of the main paper over draws of $\Deval$ is $\sum_y\bigl(\sum_gP(y,g\mid x)r_k(y,g)\bigr)\log u(y)$, maximized by $u(y)\propto\sum_gP(y,g\mid x)r_k(y,g)$. If the rollout matches the population and belief $k$ describes belief $b$, then $r_k(y,g)=P(b\mid y,g,x)$ and $\sum_gP(y,g\mid x)P(b\mid y,g,x)=P(y,b\mid x)\propto P(y\mid b,x)$. With $r\equiv1$ the target is $P(y\mid x)$, the pooled distribution.

\paragraph{The label-conditional list is exact.}
With $K=|\mathcal{G}|$, $w_k=P(g{=}k\mid x)$, $\eta_k(g)=\mathbf{1}[g=k]$ and $\bar U_k=P(\cdot\mid g{=}k,x)$, Eq.~(3) of the main paper gives $\hat P_h(y,g)=P(g\mid x)P(y\mid g,x)=P(y,g\mid x)$. Its responsibilities are $r_k(y,g)=\mathbf{1}[k=g]$, so it is also a fixed point of the answer update. With one annotation per person, the data determine only the $|\mathcal{Y}|\times|\mathcal{G}|$ matrix $P(y,g\mid x)$, and every nonnegative factorization of it scores the same under the log score in Eq.~(6) of the main paper. The cost $\lambda K(h)$ prefers the factorization with the fewest beliefs.

\paragraph{The posterior's optimum.}
Let $\pi=\pi_\theta(\cdot\mid x)$ and $f(h)=\mathbb{E}_z[\log\hat P_h(\Dobs)]$, the expected log-likelihood over the answers $z$ sampled for $h$. Since $R_q$ in Eq.~(7) of the main paper uses sampled answers and teacher-forced log-probabilities held constant, its expectation over $\hat h\sim q$ and the answers is $\mathbb{E}_q[f]-\beta\KL(q\,\|\,\pi)$. For any $q$, with $Z=\mathbb{E}_\pi[e^{f/\beta}]$ and $q^\star=\pi e^{f/\beta}/Z$,
\begin{multline}
  \mathbb{E}_q[f]-\beta\KL(q\,\|\,\pi)
  =-\beta\,\mathbb{E}_q\Bigl[\log\frac{q}{\pi e^{f/\beta}}\Bigr]\\
  =\beta\log Z-\beta\KL(q\,\|\,q^\star),
\end{multline}
which is maximized at $q=q^\star$~\citep{korbak2022rl}. At $\beta=1$, $q^\star$ is the Bayes posterior with prior $\pi$ and likelihood $e^{f}$, and the objective is at most $\log\mathbb{E}_\pi[e^{f}]$. Because $\hat P_h$ contains $\bar U_k$, a mean over sampled answers, $f$ is an expectation of a log. Jensen's inequality gives $e^{f(h)}\le\mathbb{E}_z[\hat P_h(\Dobs)]$, so at $\beta=1$ the objective is also a lower bound on $\log\mathbb{E}_{h\sim\pi,z}[\hat P_h(\Dobs)]$, the log-probability of $\Dobs$ under the policy's lists and answers. For $\beta<1$, $q^\star$ is a tempered posterior in which the likelihood is raised to the power $1/\beta$.

\paragraph{Exploration bound.}
If $L$ lists are sampled independently from $\pi_\theta(\cdot\mid x)$, the probability that at least one lies in $E_b$ is $1-(1-\pi_\theta(E_b\mid x))^L\le L\,\pi_\theta(E_b\mid x)$ by the union bound. If the $M$ annotations of $\Dobs$ are independent draws from the population, $\Dobs$ contains a holder of $b$ with probability $1-(1-\rho)^M$; for example $\rho=0.1$ and $M=4$ give $0.34$.

\paragraph{The label-free weight $\alpha$.}
Since $\log\hat P_h(y,g)=\log\hat P_h(y)+\log\hat P_h(g\mid y)$, the bracket in Eq.~(6) of the main paper equals $(1+\alpha)\log\hat P_h(y)+\log\hat P_h(g\mid y)$. Thus $\alpha$ raises the weight of the deployed label-free prediction from $1$ to $1+\alpha$ relative to the label term.

\section{Algorithm}
\label{supp:algorithm}

\begin{algorithm}[t]
  \caption{One BGPO training step for a question $x$.}
  \label{supp:alg}
  \small
  \begin{algorithmic}[1]
    \State Split annotations of $x$ into disjoint $\Dobs$, $\Deval$ with $M$ each
    \State Sample $L$ lists $h\sim\pi_{\thetaold}(\cdot\mid x)$ and $H$ lists $\hat h\sim q_{\thetaold}(\cdot\mid x,\Dobs)$; together they form $\mathcal{H}_x$
    \For{each list $h\in\mathcal{H}_x$ and each belief $k$ of $h$}
      \State Sample $n$ answers $z_{kj}\sim\pi_{\thetaold}(\cdot\mid x,h,k)$
    \EndFor
    \For{each list $h\in\mathcal{H}_x$}
      \State Parse, floor at $\delta$ and renormalize; compute $\bar U_k$ and $\hat P_h$ by Eq.~(3)
      \State $R(h)$ on $\Deval$ by Eq.~(6); $R_{\min}$ if $h$ is unparsable
      \State $r_k$ on $\Deval$ by Eq.~(4); $S(z_{kj})$ by Eq.~(5) and $A(z_{kj})$
    \EndFor
    \State $A(h)$ over $\mathcal{H}_x$
    \State $R_q(\hat h)$ on $\Dobs$ by Eq.~(7) and $A_q(\hat h)$ over the $H$ posterior lists
    \State Update $\theta$ on \cref{supp:eq:surrogate} with all triples of \cref{supp:objective}
  \end{algorithmic}
\end{algorithm}

One step is summarized in \cref{supp:alg}; equation numbers refer to the main paper. Answers for posterior lists are sampled from the same annotation-free prompt $(x,h,k)$ as answers for policy lists. An answer that cannot be parsed is left out of $\bar U_k$ and receives the lowest score in its belief group. A list with more than $K_{\max}$ beliefs, or one that fails to parse or is truncated, receives a constant reward $R_{\min}$ that lies below the reward of any parsed list. If all $n$ answers of a belief fail to parse, $\bar U_k$ is the uniform distribution over $\mathcal{Y}$.

\section{Datasets}
\label{supp:data}

\begin{table*}[t]
  \centering
  \small
  \caption{Dataset statistics. Label information is the estimated mutual information between an annotator's choice and label, averaged over training questions. It is the Miller--Madow corrected estimate in bits; for CLEVR-Belief it is exact.}
  \label{supp:tab:data}
  \resizebox{\textwidth}{!}{%
  \begin{tabular}{@{}llllrrr@{}}
    \toprule
    Dataset & Modality & Options & Label & Questions (train/dev/test) & Annotations & Label information \\
    \midrule
    OpinionQA~\citep{santurkar2023whose} & text & survey options & political party & 1,046 / 149 / 299 & 4.94M & 0.035 \\
    GlobalOpinionQA~\citep{durmus2024towards} & text & survey options & country & 853 / 123 / 238 & sampled & 0.069 \\
    DICES~\citep{aroyo2023dices} & text & Yes / No / Unsure & race & 693 / 99 / 198 & 72,103 & 0.057 \\
    ArtELingo~\citep{mohamed2022artelingo} & image & 9 emotions & annotation language & 61,935 / 3,151 / 6,300 & 1.07M & 0.563 \\
    PARA~\citep{yang2022para} & image & 9 aesthetics scores & photography experience & 25,398 / 2,822 / 3,000 & 807,586 & 0.211 \\
    CLEVR-Belief (ours) & image & objects A--D & rule label & 16,838 / 1,871 / 4,011 & sampled & 0.460 \\
    \bottomrule
  \end{tabular}}
\end{table*}

\paragraph{Selection rule.}
A label is useful only if annotators with different labels choose differently. For each dataset and each candidate label attribute, we estimate on every training question the mutual information between an annotator's choice and label, and average it over questions. We keep the attribute with the largest average. We then repeat the estimate with the labels randomly permuted among the annotators of each question, and keep the dataset only if the observed average exceeds the permuted ones at $p<0.01$. The attribute is fixed before any model is trained. The estimate is the Miller--Madow corrected mutual information, floored at zero, and we use at least 100 permutations.

\paragraph{Text datasets.}
OpinionQA~\citep{santurkar2023whose} contains 1,498 multiple-choice questions from 15 waves of the Pew American Trends Panel; we use political party as the label. The release includes individual responses, which we use directly without survey weights: the 1,494 questions with a mapped option set give 4.94M responses from Democrat, Republican and Independent respondents. GlobalOpinionQA~\citep{durmus2024towards} contains 2,556 questions from the Pew Global Attitudes surveys (2,203) and the World Values Survey (353), and releases only per-country answer percentages. An annotation therefore draws a country from the label shares and a choice from that country's distribution. We keep the 1,214 questions with at most six options that cover the eight training countries (Germany, France, Britain, Turkey, Russia, the United States, Spain and Poland), drop the ``don't know'' and ``refused'' options, and give equal shares to the countries surveyed for a question. Lebanon and Jordan are held out and appear only at test time. DICES~\citep{aroyo2023dices} contains adversarial chatbot conversations rated Yes (unsafe), No (safe) or Unsure by demographically diverse raters, with the rater's gender, race, age group and education released (and locale in DICES-990). We use DICES-990, in which each of the 990 conversations is one question with about 73 ratings, take the overall safety rating as the choice, and use the rater's race as the label.

\paragraph{Image datasets.}
ArtELingo~\citep{mohamed2022artelingo} annotates WikiArt paintings with one of nine emotion categories (amusement, awe, contentment, excitement, anger, disgust, fear, sadness, something else) and a caption, in English, Arabic and Chinese. Records carry the annotation language but no annotator identifier. The dataset uses language as a proxy for culture, and so do we. PARA~\citep{yang2022para} contains 31,220 photos, each rated by about 25 of 438 subjects. Each rating includes an aesthetics score from 1 to 5 in steps of 0.5, which gives nine options, and the released subject attributes include photographic and artistic experience. We use photographic experience, with the levels beginner, competent, proficient and expert, as the label. For both image datasets a training image has at least eight annotations.

\paragraph{CLEVR-Belief.}
We take CLEVR~\citep{johnson2017clevr} scenes with three to ten objects, whose scene graphs give each object's size, material, color and pixel coordinates. In every scene we mark three or four objects with the letters A--D; the unmarked objects are background. The question reads ``Consider only the objects marked A, B, C, or D that appear in this image. Unmarked objects are not choices. Different people may prefer different marked objects. Which marked object would a randomly chosen person pick?''. The four rules pick the large object, the metal object, the red object, and the leftmost object (smallest horizontal pixel coordinate) among the marked ones. A rule is valid in a scene if exactly one marked object satisfies it, and the leftmost object must lie at least 10 pixels to the left of the next one. Every scene has exactly three valid rules, always including red and leftmost. Marked objects must be clearly visible: the marker stays within a quarter of the object's radius from its center, and objects hidden behind others are not marked. This leaves 22,720 scenes. CLEVR training scenes give 16,838 training and 1,871 development questions, and CLEVR validation scenes give the 4,011 test questions. Each synthetic rater follows one rule and picks its object. The red rule is the rarest, with share $\rho_{\min}=0.1$ in the main paper, and the other two valid rules have share 0.45 each. A rater carries the label of its rule with probability $\kappa=0.7$ and otherwise one of the other three labels chosen uniformly. Each training question comes with eight sampled annotations, and test questions are scored against the exact distributions.

\section{Reading the tables}
\label{supp:reading}

This section explains how to read the metrics in the result tables of the main paper. Lower is better for every JSD. A method that learns only the pooled distribution can reach a low overall JSD, but its conditional and worst-group JSD stay high on questions where groups disagree. Higher is better for the recovery rate: 1 means that a method captures all the differences between labels that the annotations show, 0 means that its prediction is the same for every label, as for any method that ignores the label, and a negative value means that the label makes its prediction worse. The rate can exceed 1 when the label-free prediction is poor, so we read it together with overall JSD. On synthetic data, a low Small. shows that the rare belief was found and not absorbed into a larger one. ARI is 1 when every annotation is credited to its rater's true belief and near 0 when the credit is no better than chance.

\section{Recovery rate}
\label{supp:recovery}

\paragraph{Definition.}
For a test question $x$ with annotations $\mathcal{D}_x$, let $\rho_g$ be the share of annotations with label $g$, $q_g$ the empirical choice distribution of label $g$, and $\bar q=\sum_g\rho_g q_g$ the pooled one. A method supplies a label-conditional prediction $u_g$ and a label-free prediction $\hat p$; for BGPO these are $\hat P_h(\cdot\mid g)$ and $\hat P_h(\cdot)$, and for a method trained with labels $\hat p=\sum_g\hat P(g\mid x)\,\hat P(\cdot\mid g,x)$. The label gain and the plug-in mutual information of $x$ are
\begin{align}
  \Delta(x)&=\textstyle\sum_g\rho_g\sum_y q_g(y)\bigl[\log u_g(y)-\log\hat p(y)\bigr], \label{supp:eq:gain}\\
  \hat I(Y;G\mid x)&=\textstyle\sum_g\rho_g\,\KL(q_g\,\|\,\bar q), \label{supp:eq:mi}
\end{align}
and the recovery rate of the main paper is $\mathrm{Rec}=\sum_x\Delta(x)/\sum_x\hat I(Y;G\mid x)$. The gain $\Delta(x)$ in \cref{supp:eq:gain} is the mean of $\log u_g(y)-\log\hat p(y)$ over the annotations $(y,g)\in\mathcal{D}_x$, and $\hat I(Y;G\mid x)$ in \cref{supp:eq:mi} is the same gain at $u_g=q_g$, $\hat p=\bar q$. We sum over questions before dividing, so that questions with $\hat I(Y;G\mid x)$ near zero do not dominate.

\paragraph{Reading the value.}
Adding and subtracting the entropies of $q_g$ and $\bar q$ in \cref{supp:eq:gain} gives
\begin{equation}
  \Delta(x)=\hat I(Y;G\mid x)+\KL(\bar q\,\|\,\hat p)-\textstyle\sum_g\rho_g\,\KL(q_g\,\|\,u_g).
  \label{supp:eq:gain-decomp}
\end{equation}
A method whose prediction ignores the label has $u_g=\hat p$ for every $g$, so $\Delta(x)=0$ and $\mathrm{Rec}=0$ whatever its label-free error. The gain is largest at $u_g=q_g$, hence
\begin{equation}
  \mathrm{Rec}\le1+\frac{\sum_x\KL(\bar q\,\|\,\hat p)}{\sum_x\hat I(Y;G\mid x)},
  \label{supp:eq:rec-bound}
\end{equation}
which equals $1$ when the label-free prediction is the pooled distribution. A value above $1$ therefore reflects label-free error, which overall JSD reports, not information beyond the label; we read the two together. $\mathrm{Rec}<0$ means that the label-conditional predictions score worse on the annotations than the label-free one. The reference is the method's own label-free prediction and not $\bar q$: with $\bar q$ as reference the numerator would be $\hat I(Y;G\mid x)-\sum_g\rho_g\KL(q_g\,\|\,u_g)$, which also charges the label-free error and equals $-\KL(\bar q\,\|\,\hat p)\le0$ for every method that ignores the label.

\paragraph{Estimation.}
The plug-in $\hat I(Y;G\mid x)$ overestimates $I(Y;G\mid x)$ in finite samples. The same estimate is the denominator for every method, so it does not change their order.

\section{Hyperparameters}
\label{supp:hparams}

\begin{table}[t]
  \centering
  \small
  \caption{BGPO hyperparameters.}
  \label{supp:tab:hparams}
  \begin{tabular}{@{}lr@{}}
    \toprule
    Setting & Value \\
    \midrule
    Policy lists $L$ / posterior lists $H$ per question & 4 / 4 \\
    Answers per belief $n$ & 2 \\
    Annotations $M$ in each of $\Dobs$, $\Deval$ & 4 \\
    Maximum beliefs per list $K_{\max}$ & 4 \\
    Label-free weight $\alpha$ & 0.5 \\
    Posterior temperature $\beta$ & 0.5 \\
    Floor $\delta$ & 0.01 \\
    Cost per belief $\lambda$ & 0.02 (0 for the first 100 steps) \\
    Reward of an unparsable list $R_{\min}$ & $(2+\alpha)\log(\delta/2)-\lambda K_{\max}-1=-14.3$ \\
    Clip range $(\epsilon_{\text{low}},\epsilon_{\text{high}})$ & (0.2, 0.28) \\
    Sampling temperature & 1.0 \\
    Maximum list length & 160 tokens \\
    Maximum answer length & 192 tokens (reasoning $\le$ 160) \\
    Learning rate & $10^{-4}$, cosine decay, 20 warm-up steps \\
    Training steps / questions per step & 500 / 4 \\
    LoRA rank / scale & 16 / 32 \\
    Baseline group size / maximum response & 8 / 512 tokens \\
    Seeds & 3 (main results), 1 (sweeps) \\
    Framework & verl~\citep{sheng2025hybridflow} \\
    \bottomrule
  \end{tabular}
\end{table}

The settings are listed in \cref{supp:tab:hparams}.

\paragraph{Generation budget.} Per question and step, BGPO samples $L+H=8$ lists and up to $(L+H)K_{\max}n=64$ answers, 72 sequences in all. At the length limits this is at most $8\times160+64\times192=13{,}568$ tokens. The baselines sample $8\times512=4{,}096$ tokens. BGPO reads $2M=8$ annotations per question and step, and the baselines receive the same $2M$.

\paragraph{Reasons for the values.}
\begin{itemize}
  \item The cost $\lambda$ starts after 100 steps, so that it does not shrink lists to one belief before the answers for further beliefs are good enough to pay for it.
  \item $\delta$ bounds every log score below by $\log\delta$, which keeps rewards finite.
  \item $R_{\min}$ must lie below the lowest reward a parsed list can receive, so that unparsable lists rank last in their group.
\end{itemize}

\section{Ablation}
\label{supp:ablation}

We remove one component of BGPO at a time on CLEVR-Belief with $\kappa=0.7$ and report the results in \cref{supp:tab:ablation}. Without posterior lists, the policy is trained on lists written from the question alone. Without responsibility credit, each annotation is credited by its label alone, as in label conditioning.

\begin{table}[t]
  \centering
  \small
  \setlength{\tabcolsep}{4pt}
\caption{\textbf{Ablation on CLEVR-Belief.} Results are on the development set after 100 training steps.}
  \label{supp:tab:ablation}
    \begin{tabular}{@{}l cccc@{}}
    \toprule
    Variant & Cond.\,$\downarrow$ & Worst\,$\downarrow$ & Overall\,$\downarrow$ & Rec.\,$\uparrow$ \\
    \midrule
    \textbf{BGPO} & \textbf{0.0584} & \textbf{0.0778} & \textbf{0.0343} & \textbf{0.825} \\
    w/o posterior lists & 0.0763 & 0.0972 & 0.0427 & 0.780 \\
    w/o responsibility credit & 0.1275 & 0.1533 & 0.0397 & 0.000 \\
    \bottomrule
  \end{tabular}
\end{table}

\section{Prompts}
\label{supp:prompts}

Every prompt starts with the same question block: the sentence ``The item below was shown to many different people. Each person answered its final question by picking exactly one option.'', the question (and image), the options lettered from A, and the line ``Every person carries one of these labels:'' followed by the label set $\mathcal{G}$. Each prompt ends with one worked example for an unrelated question, in the output format requested.

\paragraph{List prompt (policy).} After the question block:
{\footnotesize
\begin{verbatim}
People pick different options because they
hold different beliefs about how the question
should be answered. Describe this population
of people as 1 to 4 distinct beliefs, one
line per belief:
Belief <number>: <one short sentence> |
  share=..% | <label 1>=..% <label 2>=..% ...
share is the percentage of all people who
hold the belief (shares sum to 100). The
label percentages say who holds the belief:
among its holders, the percentage carrying
each label (they sum to 100 on every line;
include every label).
Now write the belief lines for the item
above. Output only the belief lines.
\end{verbatim}}

\paragraph{Posterior prompt.} The list prompt, followed by the annotations of $\Dobs$:
{\footnotesize
\begin{verbatim}
A few people from this population were
observed, written as (choice, label):
(A, <label>); (B, <label>); ...
Use them as evidence about which beliefs are
present, how common each is and who holds it.
\end{verbatim}}

\paragraph{Answer prompt.} After the question block and the list $h$:
{\footnotesize
\begin{verbatim}
Consider only the people who hold belief k:
"<description of belief k>"
What percentage of them pick each option?
Holders of one belief need not all pick the
same option. Give one sentence of reasoning,
then the percentages for all options
(summing to 100):
Reasoning: <one sentence>
Answer: A=..% B=..% C=..% ...
\end{verbatim}}

\paragraph{Baseline prompts.} The baselines read the same question block. SFT and PRDO state one distribution over the options, GRPO states a single option, and the label-in-prompt variants also read the annotator's label.